\documentclass{article} % For LaTeX2e
\usepackage{iclr2027/iclr2027_conference,times}
\iclrfinalcopy

\usepackage{amsmath,amsfonts,bm}

\def\eqref#1{equation~\ref{#1}}
\def\1{\bm{1}}

\DeclareMathAlphabet{\mathsfit}{\encodingdefault}{\sfdefault}{m}{sl}
\SetMathAlphabet{\mathsfit}{bold}{\encodingdefault}{\sfdefault}{bx}{n}

\usepackage{graphicx}
\usepackage{xcolor}
\usepackage{longtable}
\usepackage{caption}

\newlength{\framewidth}
\newlength{\framesetwidth}

\newcommand{\framesetlayout}{%
    \setlength{\framewidth}{0.16\linewidth}%
    \setlength{\tabcolsep}{0.002\linewidth}%
}

\newcommand{\framecell}[2]{%
    \begin{minipage}[t]{\framewidth}
        \centering
        \includegraphics[width=\linewidth]{#1}\par
        \vspace{2pt}
        {\scriptsize
            \raisebox{2pt}[\height][\depth]{#2}\par
        }
    \end{minipage}%
}

\newcommand{\framesetheader}{%
    \begin{minipage}{\framesetwidth}
        \centering
        \framesetlayout

        \begin{tabular}{@{}c|c@{}}

            \makebox[
                \dimexpr2\framewidth+2\tabcolsep\relax
            ]{Input}%

            &

            \makebox[
                \dimexpr4\framewidth+6\tabcolsep\relax
            ]{Generated frames}%

        \end{tabular}\par
    \end{minipage}%
}

\newcommand{\frameset}[3]{%
    \begin{minipage}{\framesetwidth}
        \centering
        \framesetlayout

        {\small\color{gray}#1\par}
        \vspace{4pt}

        \begin{tabular}{@{}c|c@{}}

            \begin{tabular}[t]{@{}cc@{}}
                #2\unskip%
            \end{tabular}%

            &

            \begin{tabular}[t]{@{}cccc@{}}
                #3\unskip%
            \end{tabular}%

        \end{tabular}\par

        \vspace{12pt}
    \end{minipage}%
}

\usepackage[utf8]{inputenc} % allow utf-8 input
\usepackage[T1]{fontenc}    % use 8-bit T1 fonts
\usepackage{fix-cm}         
\usepackage{hyperref}       % hyperlinks
\usepackage{url}            % simple URL typesetting
\usepackage{booktabs}       % professional-quality tables
\usepackage{amsfonts}       % blackboard math symbols
\usepackage{nicefrac}       % compact symbols for 1/2, etc.
\usepackage{microtype}      % microtypography
\usepackage{xcolor}         % colors
\usepackage{amsmath}
\usepackage{cleveref}
\usepackage{graphicx}
\usepackage{multirow}
\usepackage{makecell}
\usepackage{adjustbox}

\usepackage{array}

\usepackage{tabularx}
\usepackage{verbatim}       %for comments
\usepackage{soul}           %for highlights

\usepackage{bm} 
\newcommand{\best}[1]{\ensuremath{\bm{#1}}}
\newcommand{\secondbest}[1]{\ensuremath{\underline{#1}}}

\title{Neural Voxel Dynamics: Learning Volumetric Feature Advection for 3D Physics in \\ V-JEPA Latent Space}

\author{Zican Wang \\
Department of Computer Science \\ 
University College London \\
UK \\
\texttt{\{zcabanj\}@ucl.ac.uk} \\
\And
Niloy Mitra \\
Department of Computer Science \\
University College London \\
UK \\
\texttt{\{n.mitra\}@ucl.ac.uk} \\
}

\begin{document}
\maketitle
\lhead{Preprint. Under review.}
\vspace{-.2in}
\begin{abstract}
We present \textit{Neural Voxel Dynamics}, a self-supervised framework for learning 3D latent dynamics from monocular video. While generative video models now produce visually compelling motion, their predominantly 2D representations provide limited geometric structure for modeling and controlling physical interactions. 
Instead, we learn dynamics in a lifted volumetric latent space. Specifically, we unproject semantic Video Joint-Embedding Predictive Architecture (V-JEPA) features into a voxel grid using monocular depth priors, producing a geometrically grounded representation that retains rich video features. 
We then introduce \textit{Volumetric Feature Advection}, an action-conditioned transition model that predicts the evolution of these latent features directly in 3D. Unlike hybrid approaches that assume access to privileged physics-engine states or explicit simulation during training and/or inference, we learn only using   video-derived supervision and, if available, actions, allowing a single latent dynamics model to represent heterogeneous material-dependent phenomena, unifying rigid-body and fluid dynamics. 
We evaluate across multiple datasets (synthetic and real) and unseen scenarios (e.g., new boundary conditions, OOD generalization), measuring both predictive fidelity and 3D geometric consistency. 
We report physical plausibility with Physics-IQ using videos decoded from predicted latent states and videos generated from predicted optical flow. 
Our results show that geometrically lifting pretrained video latent representations provides a scalable route toward dynamic world models that learn structured 3D physical dynamics without access to privileged simulator state.
\end{abstract}
\section{Introduction}

\textit{Can we learn implicit physics from videos and use it to generate physically grounded dynamic worlds?}
Modern video models learn rich representations of appearance and motion, yet typically predict dynamics in 2D image-aligned or token-based latent spaces. Physical dynamics, however, unfold in 3D: objects persist through occlusion, motion occurs in a common spatial frame, and interactions depend on geometry and material properties. While recent large-scale video models achieve impressive visual fidelity, they may struggle to maintain such physical invariants over time, particularly under interaction~\citep{motamed2026physIQ}. We humans, in contrast, regularly infer and predict physical behavior from only visual observations without explicitly solving for the parameters of a physical simulator~\citep{mccloskey1983intuitive,battaglia2013simulation}.

Existing approaches loosely trade off \textit{explicit physical structure} and \textit{learned flexibility}. Neural-physics hybrids~\citep{wangPhysCtrlGenerativePhysics2025,liWonderPlayDynamic3D2025,xiePhysGaussianPhysicsIntegrated3D2024a,tanPhysMotionPhysicsGroundedDynamics2024,zhanPerpetualWonderLongHorizonActionConditioned2026} couple learned visual representations with classical simulators. While these approaches, by construction, provide strong physical and geometric structure grounding, they require many assumptions such as predicting simulator states and physical parameters from visual observations, and selecting an appropriate material model or simulator a priori. On the other hand, learned video dynamics models~\citep{nguyen2020blockgan,ehrhardt2020relate,gillmanForcePromptingVideo2025,guDiffusionShader3Daware2025}, being implicit, can model non-trivial visual phenomena without requiring any classical simulator. Although some incorporate 3D-aware structure, most evolve state in 2D or image-aligned latent spaces; hence, 3D motion, occlusion, and interaction continue to be represented implicitly. We investigate a middle ground: \textit{3D geometry grounded, yet materially implicit} dynamics. 

Our approach leverages the rich semantic representations of Video JEPA (V-JEPA)~\citep{assranVJEPA2SelfSupervised2025,mur-labadiaVJEPA21Unlocking2026}, which we unproject into a 3D voxel grid using monocular depth priors~\citep{renGen3C3DInformedWorldConsistent2025a}. The lifting places features observed across frames into a single Euclidean coordinate system, allowing the dynamics model to reason directly about spatial motion rather than jointly representing dynamics and camera projection in an image-aligned latent space (i.e., camera motion is factored out). Monocular observations, however, reveal only a partial view of the underlying 3D world, so we additionally track latent state in unobserved voxels over time; note we do not perform object completion. Our persistent volumetric state allows information to survive occlusion to some extent and provides an easier representation for reasoning about object permanence. Our key hypothesis is that such a representation turns future prediction from a sequence-to-sequence problem into a \textit{spatio-temporal latent transport problem in 3D}.

\begin{figure}[t!]
    \centering    \includegraphics[width=\linewidth]{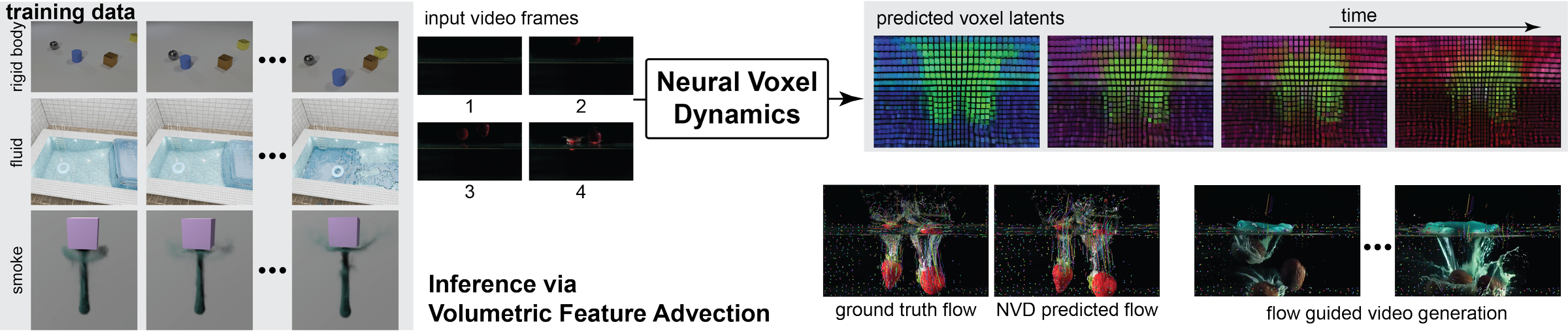}
    \caption{
    \textbf{Unified Implicit 3D Physics.} Trained exclusively on unconstrained 2D videos (left), Neural Voxel Dynamics learns to simulate complex interactions by advecting latent features within a 3D voxel grid (right). Our implicit formulation naturally unifies the prediction of heterogeneous materials like fluids, smoke, and rigid bodies. Here, conditioned on four initial frames, our model predicts, in V-JEPA space, the complex dynamics of fruits dropping into water; this is then decoded to predict 2D flows and drive a trajectory-driven video model. See supplemental for videos. 
    }
    \label{fig:teaser}
\end{figure}

To evolve this latent world state, we introduce \textit{Volumetric Feature Advection}, an action-conditioned predictor that performs grid-to-grid updates in the lifted 3D representation. Here, an \textit{action} can be specified explicitly as a 3D external force vector or implicitly using multiple past frames. 
Given the current volume and action, our predictor, modeled as a flow matching transformer, estimates a per-voxel 3D flow field and transports V-JEPA features by locally attending and aggregating the volume along it, while specifically adjusting predictions being in observed or occluded regions. 
We represent material-dependent behavior implicitly by the high-dimensional latent state and the learned transition function. 
Consequently, the same transition architecture can model different physical behaviors (e.g.,  rigid-body motion, fluid-like transport, or hybrids) without any material-specific solver. Our predictor is trained end-to-end only using video-derived targets and actions, without supervision from internal physics-engine states or manual physical annotations. \Cref{fig:teaser} visualizes predicted latent states as the model evolves a partially observed scene of fruits falling into water. 

We evaluate Neural Voxel Dynamics across heterogeneous physical domains using CLEVRER~\citep{yi2019clevrer}, PhysInOne~\citep{zhou2026physinonevisualphysicslearning}, PhysGaia~\citep{kim2026physgaia}, and Kubric~\citep{greff2022kubric}. We measure predictive fidelity (feature-space error and reconstructed-frame PSNR), geometric consistency (occupancy loss), and generalization (e.g., changing object counts, shapes, or material configurations unseen during training). We further assess the physical plausibility of generated videos using Physics-IQ~\citep{motamed2026physIQ}. Against representative neural-physics hybrids~\citep{wangPhysCtrlGenerativePhysics2025,xiePhysGaussianPhysicsIntegrated3D2024a,liuPhysGenRigidBodyPhysicsGrounded2025}, our NVD reduces feature-space prediction error by 6\% - 19\% without access to simulator state; against a 2D latent diffusion video model~\citep{yang2024cogvideox}, it improves dynamic consistency for synthetic videos (evaluated from Physics IQ benchmark) by up to 28\%. It also improves the same video models on real videos by a score of 0.2 by injecting predicted trajectory controls.
Together, these results suggest that explicit 3D structure with spatially anchored latent features is a useful inductive bias for learning heterogeneous physical dynamics.

Our contributions are: 
(i) a persistent, geometrically grounded latent world state that lifts pretrained video features into 3D, maintaining partially observed state through time; 
(ii) Volumetric Feature Advection, an action-conditioned transition operator that generatively evolves this state without material-specific solvers or privileged information; and 
(iii) empirical evaluation showing that a single shared 3D latent transition model captures diverse dynamics outperforming hybrid and 2D-latent baselines on flow and video prediction.

\section{Related Works}
\vspace{-2mm}

\paragraph{Video Generation and Controllability.}
Diffusion models are the dominant paradigm for video synthesis, extending foundational image-based frameworks such as {DDPM}~\citep{hoDenoisingDiffusionProbabilistic2020} and {DDIM}~\citep{songDenoisingDiffusionImplicit2022} to the temporal domain through spatio-temporal attention and latent diffusion architectures~\citep{hoVideoDiffusionModels2022, rombachHighResolutionImageSynthesis2022}. While existing video models and world models demonstrate remarkable scalability, they often lack explicit mechanisms for physically-grounded generation or control. 
To bridge this gap, recent research has explored trajectory-based conditioning. Multiple methods~\citep{he2024cameractrl,fu20243dtrajmaster} enable explicit camera or object trajectory transfer, while point-based motion guidance techniques~\citep{gengMotionPromptingControlling2025a,gillmanForcePromptingVideo2025,chu2025wan} allow for user-specified 2D motion cues. However, these approaches primarily operate in image space, where control signals remain heuristic. More  3D-aware representations~\citep{guDiffusionShader3Daware2025,shengFlexAMFlexibleAppearanceMotion2026,bhat2026loosecontrolvideodirectorialvideocontrol} do not explicitly model the underlying physical forces or 3D interactions, often resulting in motions that lack causal and/or dynamical consistency.

\paragraph{Explicit Physics-Based Generative Methods.}
A parallel line of work integrates classical physics simulation directly into generative pipelines. Early approaches utilized image-space modal bases~\citep{davisImagespaceModalBases2015} or 2D physics proxies like {PhysGen}~\citep{liuPhysGenRigidBodyPhysicsGrounded2025}. More recent approaches utilize 3D Gaussian Splatting as a substrate for simulation; e.g., {SpringGauss}~\citep{zhongReconstructionSimulationElastic2025} and {PhysGaussian}~\citep{xiePhysGaussianPhysicsIntegrated3D2024a} respectively incorporate spring-mass systems or Material Point Method (MPM)~\citep{stomakhin2013material, jiangMaterialPointMethod2016} to achieve physically plausible dynamics. While producing high-fidelity output, these methods require scene-specific reconstructions and explicit parameter estimation.

Hybrid methods~\citep{chenPhysGen3DCraftingMiniature2025a,tanPhysMotionPhysicsGroundedDynamics2024,lin2025phys4dgen} combine explicit simulation with generative priors, often relying on external engines like {PyBullet}~\citep{coumans2016pybullet}, {MuJoCo}~\citep{todorov2012mujoco}, or {Warp}~\citep{zong2023neural}. However, a fundamental bottleneck remains: these systems assume access to precise material properties and system states as input, which are often manually specified or crudely estimated. To mitigate this, {PhysDreamer}~\citep{zhangPhysDreamerPhysicsBasedInteraction2025}, {Physics3D}~\citep{liuPhysics3DLearningPhysical2024}, and {PSIVG}~\citep{fooPhysicalSimulatorIntheLoop2026} refine physical parameters through generative feedback. Despite such advances, the methods are restricted to preprocessing, segmentation, and isolated solid objects, and struggle to model complex multi-object interactions or diverse materials. Approaches like {PhysCtrl}~\citep{wangPhysCtrlGenerativePhysics2025} use diffusion models as neural proxies for simulators to bypass explicit tuning, yet still face challenges in robust material estimation and handling different physical interactions (e.g., fluids versus rigid body dynamics). 

\paragraph{Implicit Representations and Predictive World Models.}
Alternately, implicit generative models learn structured scene representations via latent decomposition. Early works like {BlockGAN}~\citep{nguyen2020blockgan} and {BlobGAN}~\citep{epsteinBlobGANSpatiallyDisentangled2022a} demonstrate the ability to disentangle scene components into latent variables, though they generally lack temporal dynamics or force-based reasoning. Joint-Embedding Predictive Architectures (JEPA)~\citep{assranSelfSupervisedLearningImages2023}  have shown promise in learning predictive world models. Recently, {V-JEPA}~\citep{drozdovVideoRepresentationLearning2024, assranVJEPA2SelfSupervised2025}, 3D-JEPA~\citep{hu3DJEPAJointEmbedding2024} and {LeWM}~\citep{maesLeWorldModelStableEndtoEnd2026} learn high-level representations for future state prediction. Concurrent work Phantom~\citep{shenPhantomPhysicsInfusedVideo2026}  
also looks at modeling the visual and latent physics jointly. While these models capture the \textit{gist} of physical movement, they lack any fine-grained, user-addressable force control required for 4D animation. Our work bridges this divide by providing a factorized 4D representation that can integrate geometric strength of reconstruction priors (e.g., particles, splats~\citep{tangLGMLargeMultiview2025}, deformation fields~\citep{sabathierActionMeshAnimated3D2026a}) with the generality of a disentangled motion (V-JEPA) latents.

\section{Method}
\label{sec:method}

\begin{figure}[t]
    \centering
    \includegraphics[width=1.0\linewidth]{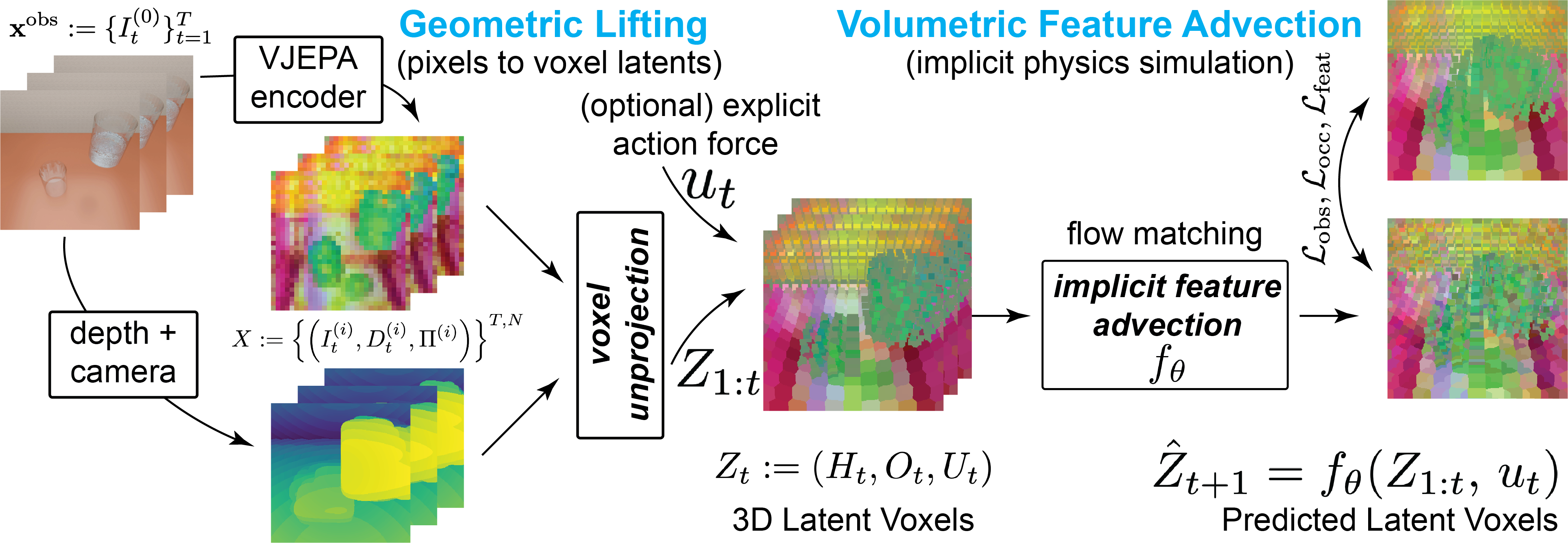}
    \caption{
    \textbf{Neural Voxel Dynamics.} By lifting 2D semantic features into a 3D latent
    voxel grid, our generative feature advector $f_\theta$ learns to implicitly
    simulate action-conditioned  physics directly in the latent space via flow
    matching; the model (comprising multiple stages of \textit{local attention} and
    \textit{temporal attention} layers, see \cref{app:model}) is supervised with
     video-derived signals and optional explicit action.
    }
    \label{fig:pipeline}
\end{figure}

Learning physics from raw video is challenging as pixel observations entangle
distinct factors: camera perspective, scene geometry, and material-dependent
dynamics. 
In contrast to 2D generative models or explicit 3D simulators, we propose \textit{Neural Voxel Dynamics} as a latent middle ground. We factorize the problem 
(\Cref{fig:pipeline}): (i)~\textit{Geometric Lifting} (\Cref{sec:lifting}), which
unprojects monocular semantic features into a partially observed 3D latent
voxel grid in the world frame; and (ii)~\textit{Volumetric Feature Advection} (\Cref{sec:vox_pred}), an
action-conditioned generative operator that evolves this spatial latent state forward in
time. Our design is: 
(i)~\emph{materially implicit:} by evolving dynamics in a high-dimensional V-JEPA latent
space, a single operator handles heterogeneous phenomena without any material label;  
(ii)~\emph{geometrically grounded:} the voxel structure makes spatial occupancy and
occlusion explicit, rather than something the model must infer from entangled 2D
latents; 
(iii)~\emph{perspective-independent:} being anchored in a world frame, it supports multiview input and a predicted 4D trajectory can be rendered from novel
viewpoints \textit{without} re-running the dynamics
operator.

\paragraph{Problem setting.}
We are given $T$ frames $\mathbf{x}^{\text{obs}} := \{I_t\}_{t=1}^{T}$ from a single
fixed viewpoint. Geometric Lifting maps these observations to a sequence of 3D latent
voxel states $Z_{1:t}$, each carrying monocular depth estimates and semantic V-JEPA
features, and marks every voxel as \texttt{empty}, \texttt{occupied}, or
\texttt{occluded}. Given a history $Z_{1:t}$ and an (optional) external action $u_t$ applied at
step $t$, the advector predicts the next state $\hat{Z}_{t+1}$; rolling this forward
autoregressively yields a full latent trajectory. Finally, a projection head projects
any predicted state back to patch-aligned latents for a target camera. For notational simplicity, we use a
single index $t$ for both observed and predicted steps; note that the V-JEPA encoder has a tubelet size of two, so one latent step spans two source frames.

\subsection{Geometric Lifting to Voxel Latents}
\label{sec:lifting}
Instead of training a 4D JEPA video encoder from scratch, we adopt a
reconstruction-based alternative that reuses a strong pretrained backbone. We
describe the monocular setting; the formulation extends directly when real
multi-view data, or a separate view-synthesis or reconstruction module, provides
additional views. We estimate per-frame depth and camera parameters with
MoGe~\citep{wang2025moge}. Given $\mathbf{x}^{\text{obs}} \in \mathbb{R}^{T \times H
\times W \times 3}$, and using $N=1$ in the monocular case,
\[
X := \left\{\left(I^{(i)}_{t}, D^{(i)}_{t}, \Pi^{(i)}\right)\right\}^{T, N}
\; \text{with } \{I,D\} \in \mathbb{R}^{N \times T \times H \times W \times (3+1)},
\]
where $I^{(i)}$ and $D^{(i)}$ are the RGB and depth observations of each view and
$\Pi^{(i)}$ its intrinsics and extrinsics; we take the first camera frame as the
world frame. In parallel, the frozen V-JEPA~2.1
encoder~\citep{mur-labadiaVJEPA21Unlocking2026} produces patch-level features, 
\[
\left\{z_{t}^{(i)}\right\}^{T'} := \mathrm{VJEPA}\!\left(\left\{I_t^{(i)}\right\}^T\right)
\in \mathbb{R}^{T' \times H' \times W' \times d},
\]
with temporal ratio $T/T' = 2$ and $d = 1408$. We use V-JEPA as it has temporally consistent dense features and has been shown to
encode a degree of physical understanding~\citep{garridoIntuitivePhysicsUnderstanding2025a},
both of which we exploit when evolving state directly in feature space.

\subsubsection{Voxel unprojection}
\label{sec: vox proj}
For each view, the frozen encoder produces a dense 2D grid of patch-level features,
and we pool the depth map to the patch resolution. Using the intrinsics, extrinsics,
and pooled depth, we unproject each patch feature $z_t$ from image space to a 3D point in
the world frame, and aggregate its contribution into a shared voxel lattice by
trilinear interpolation, yielding the feature channel $H_t$ (see \cref{app:vox-interp}).
Alongside the features we maintain two scalar channels: an \texttt{occupancy} channel
$O_t$, marking voxels that contain observed scene content, and an \texttt{occlusion} channel
$U_t$, marking cells that lie behind an observed surface and are therefore currently
unseen (i.e., a subset of the complement of $O_t$, at depths beyond the observed
surface). The full latent state is $Z_t := (H_t, O_t, U_t)$. Because only
\texttt{occupied} and \texttt{occluded} cells carry information, isolated \texttt{empty}
voxels are never materialized, saving space.

\subsection{Volumetric Feature Advection}
\label{sec:vox_pred}
Given a history $Z_{1:t}$ and action $u_t$, the advector $f_\theta$ predicts the next
state,
\[
\hat{Z}_{t+1} = f_{\theta}\!\left(Z_{1:t},\, u_{t}\right).
\]
Rather than deterministically regressing the next state, we adopt a flow-matching
formulation~\citep{lipman2022flow} (\cref{app:flow-matching}): $f_\theta$ learns a
time-dependent velocity field that transports a noisy source distribution toward the
target latent manifold. Our model has a flexible generative prior over
physically plausible transitions and lets it represent the conditional uncertainty
arising from partial observation and unresolved dynamics, without the rigid
constraints of a predefined analytical solver.

\paragraph{Explicit control.}
Our model also takes explicit (but optional) force control input $u_t$ in addition to implicit temporal frames. 
We augment each voxel token with 3D spatial and temporal positional embeddings. The
control is a time-varying action descriptor $u_t \in \mathbb{R}^{m}$: a 12-dimensional ($m=12$)
encoding of the contact point, applied direction, magnitude, temporal duration,
active-frame indicator, and local/global flags, expressed and normalized in world
coordinates to reduce ambiguity. We inject it through two complementary paths:
(i)~\textit{local integration}, in which the force is projected to the token
dimension and added element-wise to each voxel token, giving a spatially uniform
contextual field; and (ii)~\textit{global modulation}, in which the force is encoded
into a set of conditioning tokens concatenated to the sequence, letting
self-attention modulate the global scene transition (\cref{app:implementation}). Note that $u_t$ is not a privileged simulator state; it does not specify the material parameters or internal solver variables that hybrid methods require. Note that the advector never receives material parameters (e.g., stiffness, viscosity), engine states, or per-particle states; 
material-dependent behavior emerges purely from the latent state and learned transition. In our experiments, actions are drawn only for the CLEVRER/MuJoCo dataset; on data without such annotations, the action reduces to
the zero vector, so the model still predicts implicit dynamics. At inference time, users can optionally provide explicit force input (see supplemental for examples).

\paragraph{Sparse tokenization and efficiency.}
To avoid the cubic cost of dense grids, $f_\theta$ operates on a \textit{sparse set of
active voxels} and their neighbors. We define the set as the
union of \texttt{occupied} and \texttt{occluded} voxels across the context window,
dilated morphologically to leave room for predicted motion, with size $M$. This concentrates
capacity on scene-relevant geometry, and scales more with object complexity than with grid resolution. We operate at a single grid spacing. 

\paragraph{Dynamic local spatiotemporal transformer.}
Our feature advector is a Diffusion Transformer~(DiT) variant whose layers alternate between
two attention modes. (i)~\textit{Local spatial attention}, inspired by the locality
of interactions in classical simulators (e.g., MPM~\citep{jiangMaterialPointMethod2016,
welschinger2025learning}), restricts each voxel to a $k \times k \times k$ sparse
neighborhood; this center-query mechanism mimics step-wise force propagation in the physics engine and gives cost
that scales linearly with $M$, $O(Mk^3)$, rather than quadratically in the number of active
tokens. We use $5$ layers and $k=5$, with $M$ set by object complexity (i.e., number of active voxels).
(ii)~\textit{Temporal attention} regroups tokens by spatial location so that each
voxel attends causally along its own trajectory, capturing temporal changes in local features and occupancy. Because a single
step propagates information only within a local neighborhood, longer-range transport
is composed over stacked layers and successive prediction steps; we discuss the
resulting displacement range in \cref{app:model}. The transformer is modulated by the
flow-matching time $\tau$ via adaptive layer normalization
(AdaLN)~\citep{wangPhysCtrlGenerativePhysics2025,yang2024cogvideox}, outputting
feature velocities that are scattered back into the lattice; we integrate the
predicted velocity field $v_\theta$ with Euler steps to solve the probability-flow ODE.

\subsubsection{Training Objectives}
\label{sec:training}
We optimize a multi-task objective that decouples feature dynamics from geometric
occupancy across the three channels of $Z_t := (H_t, O_t, U_t)$, where $O_t$ is the
\texttt{occupancy} probability and $U_t$ the \texttt{occlusion} probability.

\paragraph{Feature velocity loss.}
Under the flow-matching formulation, the feature branch regresses a velocity field
$v_\theta$ toward the target manifold. To prevent empty space from dominating, we
weight the loss by occupancy as:
\begin{equation}
\mathcal{L}_{\mathrm{feat},r} :=
\frac{\sum_{j} O_{r}(j)\, \| v_{\theta,r}(j) - v^\star(j) \|_2^2}
     {\sum_j O_{r}(j) + \xi},
\end{equation}
where the target velocity $v^\star = \epsilon - Z_{t+1}$ is the derivative of the
linear interpolant $(1-\tau) Z_{t+1} + \tau \epsilon$ with $\epsilon$ Gaussian noise
(\cref{app:flow-matching}). The occupancy $\mathcal{L}_{O,r}$ and occlusion
$\mathcal{L}_{U,r}$ channels use the same flow-matching objective, which we find
outperforms a binary target (\Cref{tab:regression_flow_matching}), likely because it
lets geometry and features share one denoising dynamics.

\paragraph{Diffusion-forcing objective.}
Following diffusion forcing, we corrupt history frames at varying noise levels and explicitly condition the predictor on the noise level of each frame, limiting error accumulation over long
rollouts~\citep{chenDiffusionForcingNexttoken2024}. For stable autoregressive rollout, we sample an independent noise level
$\tau_r \sim p_\tau$ per latent frame~\citep{chenDiffusionForcingNexttoken2024},
shared across channels within a frame, and averaged over the clip. Each training window contains $C$ context states and
$S$ predicted states, giving $K=C+S$ states in total.
The frame index $r=C+1,\ldots,K$ covers only the
predicted states.
\begin{equation}
\mathcal L_{\mathrm{DF}}
=
\mathbb E\!\left[
\frac{1}{S}\sum_{r=C+1}^{K}
\left(
\mathcal L_{\mathrm{feat},r}
+\lambda_O\mathcal L_{O,r}
+\lambda_U\mathcal L_{U,r}
\right)
\right].
\label{eq:joint_df_objective}
\end{equation}
where the expectation is over training clips, sampled noise levels, and Gaussian
noise, and $\lambda_O, \lambda_U > 0$ weight the geometry channels.

\paragraph{Projection loss (2D).}
We encourage the projected features to match the encoded future observations in V-JEPA space. Using differentiable front-to-back compositing, we supervise the projected features and accumulated opacity of the estimated clean endpoint. For each 2D latent patch at $u,v$:
\begin{equation}
\mathcal L_{\mathrm{featproj}} =
\frac{\sum_{u,v}M_{u,v} \|\hat z_{u,v}-z_{u,v}\|_2} {D(\sum_{u,v}M_{u,v}+\delta)},
\qquad
\mathcal L_{\mathrm{occproj}} = 
\frac{1}{|\mathcal P|} \sum_{u,v}(\hat M_{u,v}-M_{u,v})^2.
\end{equation}
where $\hat{z}_{u,v}$ is the projected 2D features, $z_{u,v}$ the encoder’s target patch features, $M_{u,v}\in[0,1]$ is the reference depth-validity mask, restricted to observations inside the voxel volume, and $\hat M_{u,v}$ is the accumulated ray opacity; details in \Cref{app:softrast}. 

\paragraph{Total objective.}
The full loss combines the volumetric and projection terms,
\begin{equation}
\mathcal L=\mathcal L_{\mathrm{DF}}
+\mathbb E\!\left[
\frac{1}{S}\sum_{r=C+1}^{K}
\left(
\lambda_{\mathrm{fp}}\mathcal L_{\mathrm{featproj},r}
+\lambda_{\mathrm{op}}\mathcal L_{\mathrm{occproj},r}
\right)
\right].
\end{equation}
with weights $\lambda_{\mathrm{fp}}, \lambda_{\mathrm{op}}$ (set to $1$ in our tests). We train the voxel predictor using video-derived feature and geometry targets, conditioned on supplied force inputs, with no
supervision from physics-engine states or manual annotations.
 
\paragraph{Inference.}
At test time, we maintain a rolling history of 2 latent states (i.e., 4 image frames). Given $Z_{1:2}$
and an (optional) action $u_2$, we integrate the learned velocity field from Gaussian noise to a
clean prediction by solving the probability-flow ODE with 20 Euler steps, scatter
the result into the sparse lattice to form $\hat{Z}_{3:10}$ (i.e., 16 RGB frames), append it to the history,
and repeat. Diffusion forcing supports history-guided sampling~\citep{songHistoryGuidedVideoDiffusion2025}.
At each denoising step, we subtract the velocity predicted with fully noised history from that predicted with clean history, then add a scaled version of this difference to the history-conditioned velocity as a correction. 
\section{Experiments}
\label{sec:experiments}

\paragraph{Datasets.}
We evaluate on four sources with complementary dynamics. Our synthetic CLEVRER-style
data~\citep{yi2019clevrer} focuses on rigid-body interactions simulated with
MuJoCo~\citep{todorov2012mujoco}, augmented with a labeled force vector not present in
the original CLEVRER. PhysInOne~\citep{zhou2026physinonevisualphysicslearning} adds
fluids and rigid bodies, and PhysGaia~\citep{kim2026physgaia} contributes more
complex scenes with cloth, smoke, and soft bodies. Although the pipeline supports multi-camera inputs, to have a reasonable comparison against monocular baselines, training uses single-camera inputs. See \Cref{app:data}. 
To test \textit{zero-shot generalization}, we additionally
evaluate the \emph{frozen} NVD predictor on 30 samples from the unseen Kubric
MOVI-C dataset~\citep{greff2022kubric}. We also take the Physics-IQ~\citep{motamed2026physIQ} benchmark dataset for video physics evaluation.

\paragraph{Model variants.}
\textit{Ours-GT} uses ground-truth depth (available only for the simulated CLEVRER
data and PhysInOne) and serves as a reference for ablation; \textit{Ours-Estimate} uses
MoGe-estimated depth~\citep{wang2025moge} on all datasets and is our deployable
model for real scenes and generalization. To isolate the contribution of the 3D lift, we also train a \textit{2D baseline} of comparable capacity that removes
the 3D projection and applies the same local/temporal attention and autoregressive flow-matching objective directly on the 2D V-JEPA latents. As this baseline is
optimized with identical latent objective, it also bounds how much the V-JEPA-space metric could favor our method.

\paragraph{Baselines.}
We compare against simulator-based methods
PhysGen(2D)~\citep{liuPhysGenRigidBodyPhysicsGrounded2025} and
PhysGaussian(3D)~\citep{xiePhysGaussianPhysicsIntegrated3D2024a}, the simulator-proxy
PhysCtrl~\citep{wangPhysCtrlGenerativePhysics2025}, and the I2V model
CogVideoX~\citep{yang2024cogvideox}. All video baselines receive the same monocular initial
frames and predict the subsequent 16 frames, while PhysGaussian consumes all camera angles. PhysCtrl and PhysGaussian consume the 3D force directly; for PhysGen we project the force into 2D camera space, and for CogVideoX we supply it via text conditioning. See supplemental webpage. 

\subsection{Latent-State Prediction Across Simulation Regimes}
\label{sec:exp-latent}

We first test prediction quality directly in latent space. For each method, we compare predicted future latents to ground-truth future latents, reporting L2 error in both the 2D latent space and the 3D (unprojected from estimated depth) latent space; baseline outputs are re-encoded into V-JEPA space for a common comparison. This measures whether a model's transition stays on the V-JEPA manifold and follows the correct motion, rather than merely producing smooth-looking frames. In the multi-camera setting, all models receive only a single camera at inference but are scored across all views; in contrast, PhysGaussian is given multiple views to enable its 3DGS reconstruction. Results grouped by material appear in \Cref{tab:latent_dynamics};
the full per-dataset breakdown is in \Cref{app:exp-latent}, \Cref{tab:latent_protocols}.

\begin{table}[h]
\caption{Latent prediction on different dynamics categories (synthetic). Each entry
reports 2D / 3D latent L2 loss ($\downarrow$). Ours, being material implicit, has a single model across settings. }
\centering
\scriptsize
\begin{adjustbox}{width=1.0\textwidth, center}
\setlength{\tabcolsep}{5pt}
\begin{tabular}{lccc}
\toprule
Method & Rigid body & Fluid & Smoke \\
\midrule
CogVideoX~\citep{yang2024cogvideox}      & 3.39 $\pm$ 0.64 / 0.91 $\pm$ 0.33 & 4.21 $\pm$ 0.31 / 1.15 $\pm$ 0.13 & 3.07 $\pm$ 0.68 / 1.56 $\pm$ 0.48 \\
PhysGen~\citep{liuPhysGenRigidBodyPhysicsGrounded2025}        & 2.54 $\pm$ 0.35 / 0.59 $\pm$ 0.37 & 5.22 $\pm$ 1.10 / 1.22 $\pm$ 0.39 & 4.37 $\pm$ 0.45 / 1.80 $\pm$ 0.47 \\
PhysGaussian~\citep{xiePhysGaussianPhysicsIntegrated3D2024a}   & 3.32 $\pm$ 0.15 / 0.90 $\pm$ 0.31 & 4.80 $\pm$ 0.20 / 1.18 $\pm$ 0.19 & 2.67 $\pm$ 1.06 / 1.70 $\pm$ 0.47 \\
PhysCtrl~\citep{wangPhysCtrlGenerativePhysics2025}       & 3.58 $\pm$ 0.37 / 0.83 $\pm$ 0.33 & 5.03 $\pm$ 0.34 / 1.19 $\pm$ 0.22  & 3.72 $\pm$ 0.96 / 1.73 $\pm$ 0.48 \\
2D baseline  & 2.09 $\pm$ 0.13 / 0.31 $\pm$ 0.21 & \secondbest{2.99 \pm 0.27} / 1.20 $\pm$ 0.33 & 2.51 $\pm$ 0.53 / 1.62 $\pm$ 0.25  \\
Ours-GT        & \best{1.94 \pm 0.14} / \best{0.19 \pm 0.11} & \best{2.12 \pm 0.11} / \secondbest{0.82 \pm 0.09} & \secondbest{2.22 \pm 0.33} / \secondbest{1.15 \pm 0.20} \\
Ours-estimate  & \secondbest{1.97 \pm 0.16} / \secondbest{0.26 \pm 0.12} & \best{2.12 \pm 0.21} / \best{0.80 \pm 0.13} & \best{2.03 \pm 0.41} / \best{1.00 \pm 0.29}  \\
\bottomrule
\end{tabular}
\label{tab:latent_dynamics}
\end{adjustbox}
\end{table}

\textit{An accurate and material-robust model.}
Both our variants achieve the best or second-best 3D latent error in every material category, improving over the strongest simulator-based baseline by 7\% to up to 29\% across material types. More importantly for our central claim, our error is the \emph{most stable across materials}: the 2D latent
error varies by only $\pm 0.12$ across rigid/fluid/smoke for Ours-Estimate, whereas the specialized baselines swing sharply. PhysGen is competitive on rigid bodies but degrades markedly on fluid and smoke, reflecting their reliance on 2D structure and per-material solver assumptions. While CogVideoX catches up in the fluid and smoke scenarios, it fails in rigid-body scenarios because it lacks fine-grained control input and object permanence. PhysGaussian is primarily designed for soft-body dynamics and relies on multiview input for accurate scene reconstruction. Like PhysCtrl, it also requires segmentation of the main object and is best suited to single-object scenarios, which may partly explain its relatively weaker performance. 

Using a similar (latent) error metric as our objective did give us an advantage, as shown in the 2D baseline result being better than most other baselines. However, Ours-Estimate reduces the 3D latent error by around 29\% relative to the 2D baseline and also performs better in the additional evaluations later on. This suggests that the improvement is not solely due to the choice of metric, but also benefits from lifting into 3D. Finally, the Ours-GT ablation indicates that our method has a lighter assumption on privileged geometry.

\subsection{Flow Estimation From Latent Dynamics}
\label{sec:exp-flow}

We next test whether the learned latents encode transferable motion prediction. Keeping V-JEPA
frozen, we train a small convolutional head that maps two consecutive latent maps to
an optical-flow field (\Cref{app:flow_decoder});
predicted displacements are scaled to pixels and integrated over time. At inference, the head needs only the
latent sequence; baseline flows are computed from their generated videos.
\Cref{tab:optical_flow} reports optical-flow displacement error, normalized by the ground-truth flow magnitude, and optical flow vector angle difference. 

\begin{table*}[h!]
\centering
\caption{Optical-flow error ($\downarrow$). The table reports magnitude/angle. See supplemental for videos.}
\label{tab:optical_flow}
\scriptsize
\begin{adjustbox}{width=1.0\textwidth}
\begin{tabular}{lcccc}
\toprule
Method & CLEVRER $\downarrow$ & PhysGaia $\downarrow$ & PhysInOne $\downarrow$ & Kubric MOVI-C (zero-shot) $\downarrow$ \\
\midrule
CogVideoX      & $1.551 \pm 0.915$ / \secondbest{17.7 \pm 5.3}  & $\secondbest{0.996 \pm 0.037}$ / $\best{21.8 \pm 6.3}$ & $1.094 \pm 0.113$ / $\best{24.0 \pm 6.0}$ & $1.086 \pm 0.228$ / $\secondbest{23.4 \pm 2.7}$ \\
PhysGen        & $0.993 \pm 0.057$ / $19.2 \pm 3.6$ & $1.001 \pm 0.003$ / $50.2 \pm 7.5$ & $1.072 \pm 0.151$ / $52.3 \pm 12.9$ & $1.000 \pm 0.061$ / $24.1 \pm 20.2$\\
PhysGaussian   & $0.804 \pm 0.338$ / $26.4 \pm 4.1$ & $1.000 \pm 0.002$ / $37.8 \pm 4.0$ & $1.068 \pm 0.201$ / $60.9 \mp 13.1$ & $0.982 \pm 0.076$ / $29.7 \pm 13.8$ \\
PhysCtrl       & $10.93 \pm 20.97$ / $64.5 \pm 16.9$ & $1.085 \pm 0.173$ / $41.0 \pm 8.5$ & $1.295 \pm 0.455$ / $47.7 \pm 13.7$ & $1.047 \pm 0.121$ / $42.3 \pm 14.1$ \\
2D baseline    & $\secondbest{0.583 \pm 0.032}$ / $21.24 \pm 5.2$ & $1.055 \pm 0.124$ / $30.35 \pm 9.5$ & $\secondbest{1.005 \pm 0.314}$ / $43.33 \pm 13.2$ & $\secondbest{0.981 \pm 0.030}$ / $29.24 \pm 19.9$ \\
Ours-Estimate & $\best{0.545 \pm 0.186}$ / \best{13.5 \pm 2.5} & $\best{0.977 \pm 0.165}$ / $\secondbest{29.0 \pm 6.0}$ & $\best{0.980 \pm 0.035}$ / $\secondbest{25.9 \pm 8.6}$ & $\best{0.917 \pm 0.059}$ / $\best{23.1 \pm 9.7}$ \\
\bottomrule
\end{tabular}
\end{adjustbox}
\end{table*}

\textit{Generalizable flow across multiple datasets.}
Ours-Estimate attains the lowest flow error on all four datasets. The result holds across materials and, most notably, on unseen Kubric MOVI-C, where we improve over the best baseline
by 6.6\% in magnitude and 1.2\% in direction despite never training on it, which is evidence that the 3D-grounded dynamics generalize beyond the training distribution rather than memorizing per-dataset motion. A complementary voxel-space latent flow-consistency evaluation is deferred to \Cref{app:flow}, where Ours-Estimate also achieves the lowest error.

\subsection{Physical Plausibility (Physics-IQ)}
\label{sec:exp-physiq}
Using our generated trajectory, we generate motion-guided video with Wan-move~\citep{chu2025wan}, given the first frame and the predicted flow trajectories (see above and \Cref{app:motion_guided_video}).
We assess whether our generated dynamics are physically plausible using Physics-IQ metrics~\citep{motamed2026physIQ}. \Cref{tab:physiq_method} reports the overall
Physics-IQ verified score on synthetic videos across our four domains, in addition to a subset (10\%) of the Physics-IQ test set. And \Cref{fig:realvid_compare_main} shows a qualitative result (more in the supplementary).

\begin{table}[h]
\centering
\caption{Physics-IQ benchmark on different video datasets (overall score $\uparrow$).}
\label{tab:physiq_method}
\scriptsize
\begin{tabular}{lcccccc}
\toprule
Method & CLEVRER & Kubric & PhysGaia & PhysInOne & Physics IQ & Total \\
\midrule
CogVideoX     & 0.008 & 0.308 & \secondbest{0.296} & 0.161 & \best{0.405} & 0.237\\
PhysGen       & 0.073 & 0.126 & 0.031 & 0.194 & 0.191 & 0.163 \\
PhysGaussian (all cameras)  & \secondbest{0.368} & 0.292 & 0.023 & 0.225 & 0.253 & 0.272 \\
PhysCtrl      & 0.047 & 0.268 & 0.270 & \secondbest{0.229} & 0.289 & \secondbest{0.281} \\
Wan-move (no tracks) & 0.114 & \secondbest{0.335} & \best{0.303} & 0.226 & 0.389 & 0.259 \\
Ours-Estimate & $\best{0.478}$ & $\best{0.420}$ & 0.271 & $\best{0.293}$ & $\secondbest{0.391}$ & \best{0.384} \\
\bottomrule
\end{tabular}
\end{table}

\textit{Results.}
Ours-Estimate obtains the highest score overall and improves the score of the Wan-move baseline with no trajectories. However, performance degrades beyond 3 seconds due to latent noise, while Physics-IQ evaluates 5-second videos. Consequently, CogVideoX performs slightly better on some metrics. Full per-metric breakdowns appear in \Cref{app:physicsiq-full}.

\begin{figure}[t]
    \centering

    \setlength{\framesetwidth}{\linewidth}
    \setlength{\parskip}{0pt}

    \renewcommand{\framesetlayout}{%
        \setlength{\tabcolsep}{0.5pt}%

        \setlength{\framewidth}{%
            \dimexpr
                (\linewidth-10\tabcolsep-\arrayrulewidth)/6
            \relax
        }%
    }

    \renewcommand{\framecell}[2]{%
        \vbox{%
            \offinterlineskip
            \hbox{%
                \includegraphics[width=\framewidth]{#1}%
            }%
            \kern0.5pt
            \hbox to \framewidth{%
                \hfil{\scriptsize\strut #2}\hfil
            }%
        }%
    }

    \renewcommand{\frameset}[3]{%
        \begin{minipage}{\framesetwidth}
            \centering
            \framesetlayout

            {\scriptsize\color{gray}#1\par}
            \vspace{1pt}

            \begin{tabular}{@{}c|c@{}}

                \begin{tabular}[t]{@{}cc@{}}
                    #2\unskip%
                \end{tabular}%

                &

                \begin{tabular}[t]{@{}cccc@{}}
                    #3\unskip%
                \end{tabular}%

            \end{tabular}\par
        \end{minipage}%
    }

    {\small\framesetheader\par}
    \vspace{1pt}

    % ==================================================
    % SET 1: FRUITS DROP IN WATER
    % ==================================================
    \frameset{Wan-move guided by Our-estimate: potato in water, the 4 input frames are on the left}
    {

        \framecell{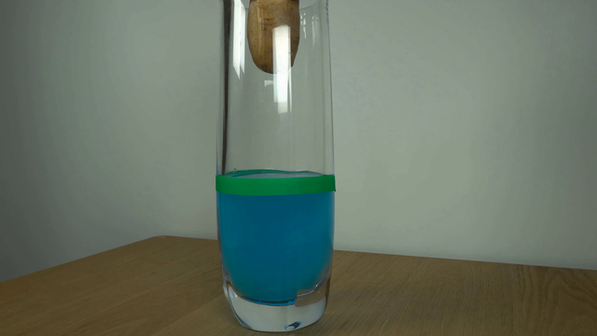}{1} &
        \framecell{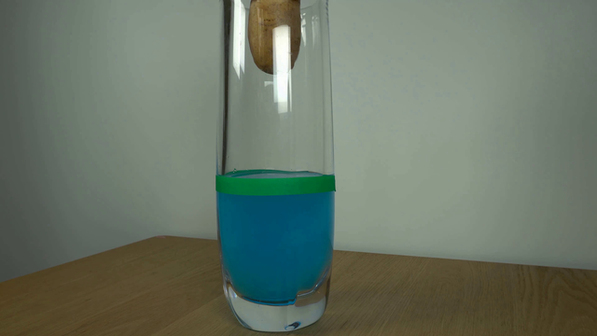}{2} \\[1pt]

        \framecell{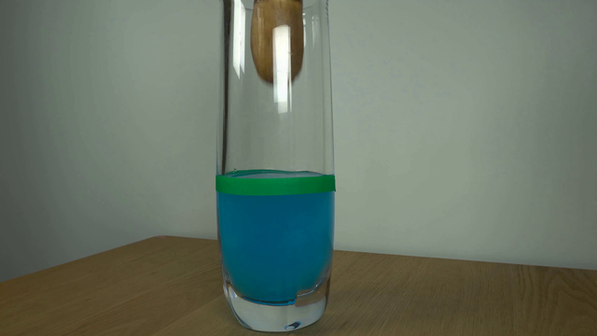}{3} &
        \framecell{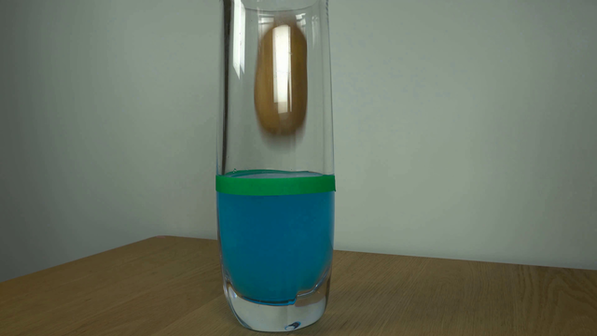}{4}
    }
    {

        \framecell{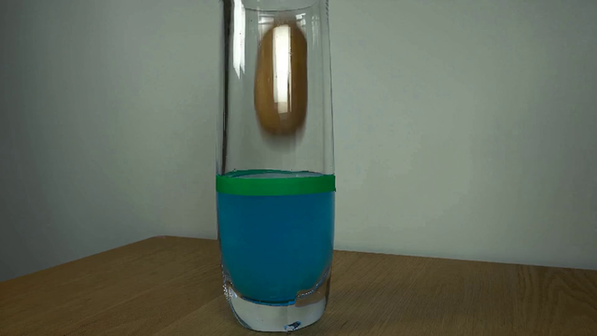}{5} &
        \framecell{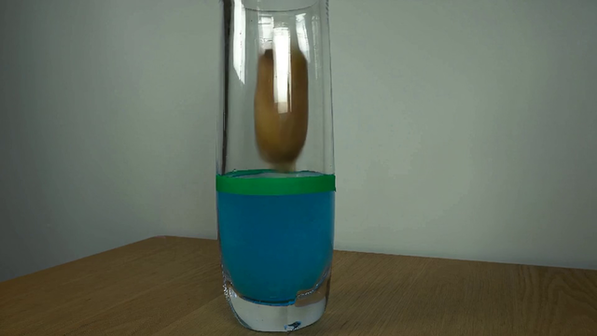}{6} &
        \framecell{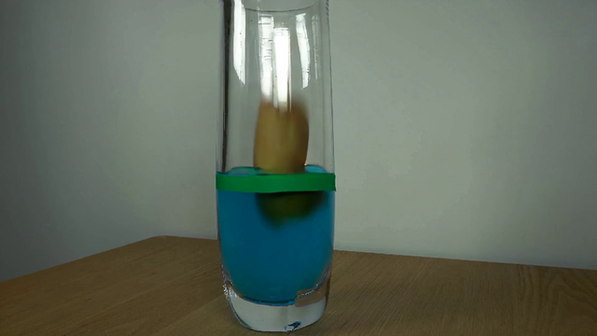}{7} &
        \framecell{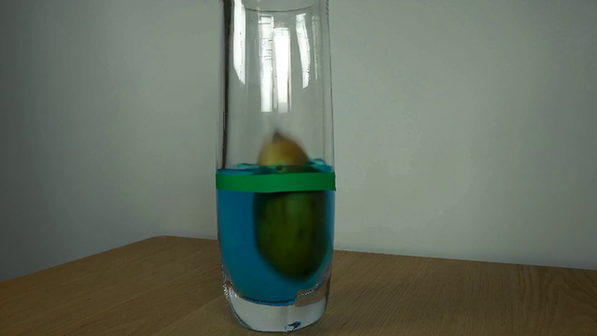}{8} \\[1pt]

        \framecell{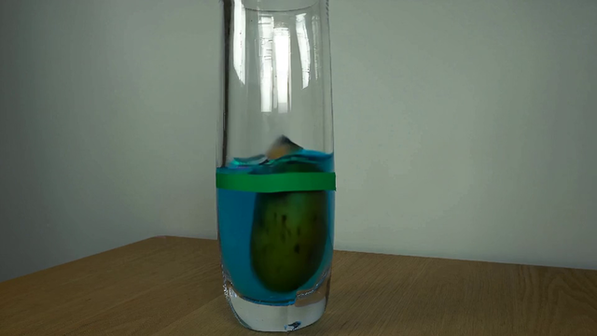}{9} &
        \framecell{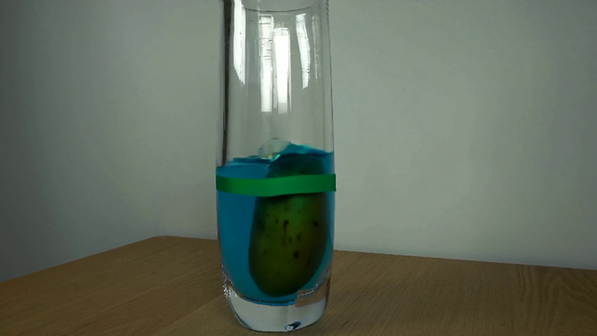}{10} &
        \framecell{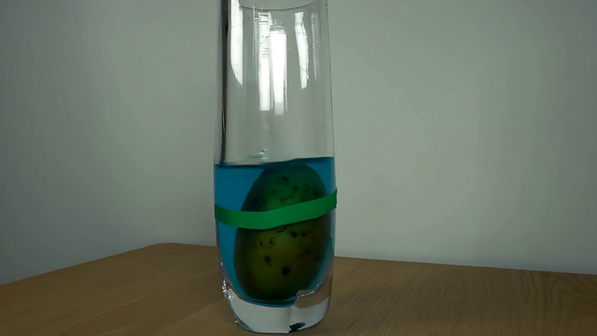}{11} &
        \framecell{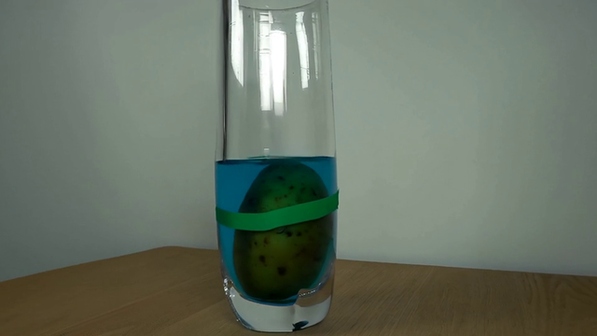}{12}
    }\par

    \vspace{3pt}

    % ==================================================
% SET 2: COFFEE POURING
% One row only, 6 columns, no input/generated split.
% ==================================================
\begin{minipage}{\framesetwidth}
    \centering
    \framesetlayout

    {\scriptsize\color{gray}CogVideoX comparison, with single input frame (frame 1) already shown above\par}
    \vspace{1pt}

    \begin{tabular}{@{}cccccc@{}}
        \framecell{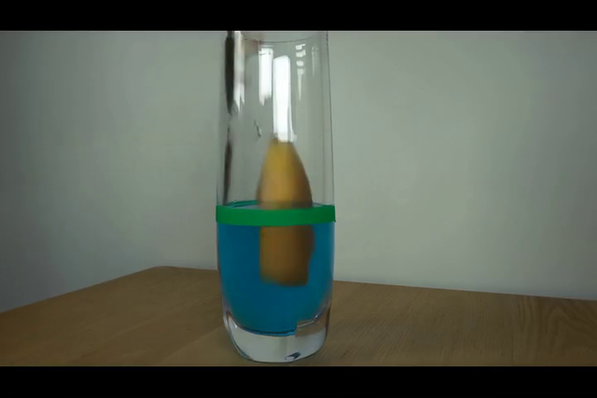}{2} &
        \framecell{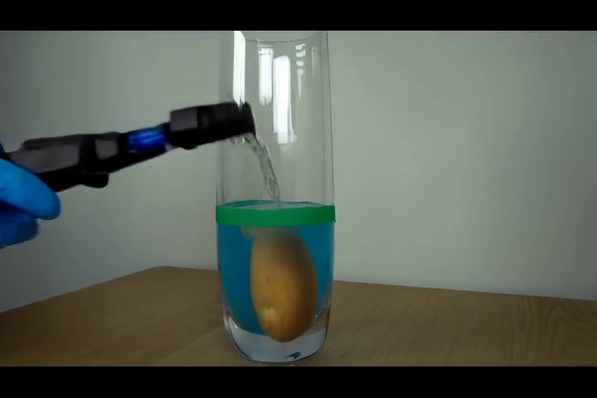}{4} &
        \framecell{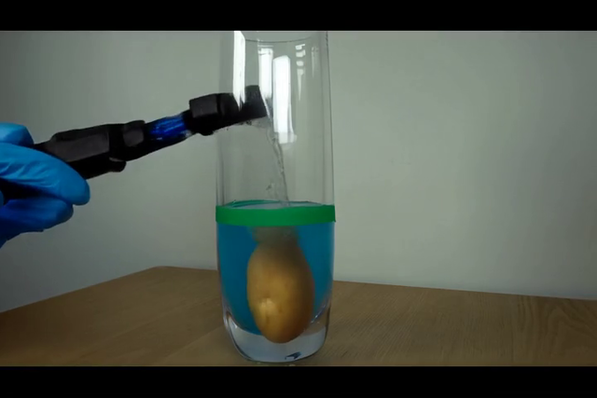}{6} &
        \framecell{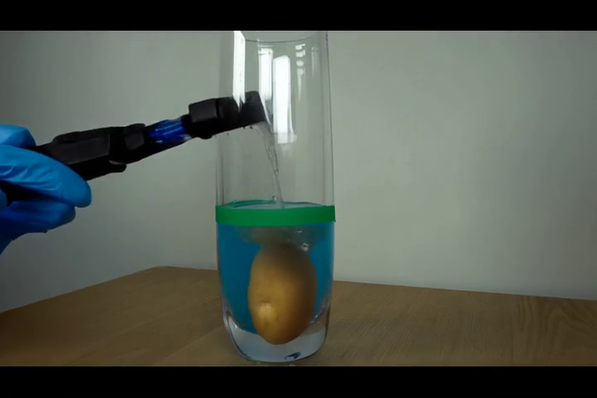}{8} &
        \framecell{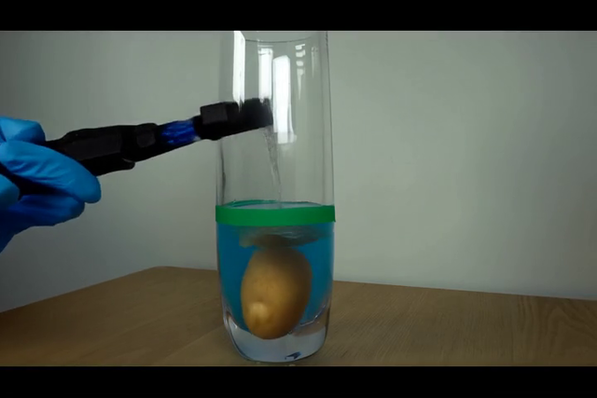}{10} &
        \framecell{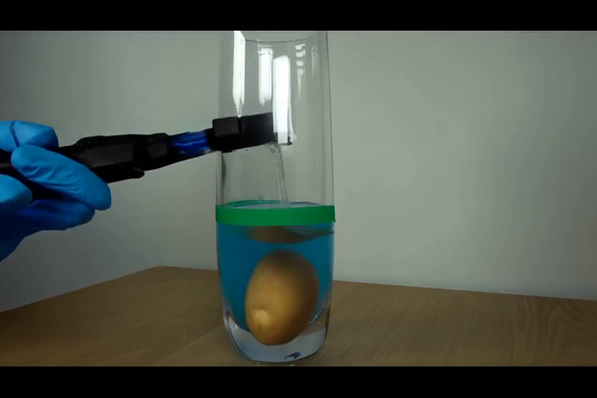}{12}
    \end{tabular}\par
\end{minipage}
    \setlength{\abovecaptionskip}{1pt}
    \caption{Input frames and  the corresponding generated sequences. Note, for ours how the water level rises as the potato sinks demonstrating materially implicit dynamics. }
    \label{fig:realvid_compare_main}

\end{figure}
\subsection{Other evaluations}
\vspace{-2mm}
\paragraph{Occupancy as a proxy for mass conservation.}
Because our transition operator predicts an explicit 3D occupancy field, we can measure how the occupancy matches the ground truth as a scene "mass" conservation. We report occupancy IoU against ground truth and the drift in total occupied volume. Full results are in \Cref{app:occupancy}. Our model maintains high occupancy IoU, consistent with the behavior seen in the flow and Physics-IQ evaluations. 
\vspace{-5mm}
\paragraph{Controllable Generalization.}
We probe controllability in a simple, interpretable setting: a single ball on a floor, actuated by an impulse force applied at chosen contact points. 
The results (see supplementary for videos) show that varying the force produces correspondingly different, physically consistent trajectories, and that the model
respects contact geometry (e.g., object permanence across temporary occlusion, plausible collision with different objects). This demonstrates that control transfers to configurations outside the training clips.
\vspace{-2mm}
\subsection{Ablations}
\vspace{-2mm}
\label{sec:exp-ablation}
We highlight the findings from our ablations here; further details are in \Cref{app:ablations}. \\
\textbf{(i) 2D vs.\ 3D latent dynamics.}
Across both latent prediction (\Cref{tab:latent_dynamics}) and flow
(\Cref{tab:optical_flow}), the 2D baseline is consistently beaten by our 3D model under the same objective and comparable capacity.
The gap widens on non-rigid and occluded content, where reasoning in a shared Euclidean frame matters most. This suggests that the improvement mainly comes from the 3D lifting rather than differences in the training setup.
\\
\textbf{(ii) Regression vs.\ generative transition.}
\Cref{app:ablations} compares a deterministic-regression variant
against the generative flow-matching formulation (with diffusion forcing and history
guidance) over 24-frame rollouts. Regression is clearly worse on occupancy IoU and
both flow metrics (0.208 vs.\ 0.709 IoU), confirming that a generative prior is
needed to model the conditional uncertainty of partially observed dynamics. Among
generative variants, vanilla history guidance (DF+HG) is best on all three metrics
and is used for Ours-Estimate.

\vspace{-4mm}
\section{Conclusion}
\label{sec:conclusion}
\vspace{-4mm}
Neural Voxel Dynamics bridges 2D video generation and 3D physics simulation. By lifting monocular video into a 3D latent voxels and learning an action-conditioned feature advection model, we achieve materially implicit, geometrically grounded physical transitions from monocular video, disentangling camera from material dynamics and forecasting heterogeneous phenomena without segmentation, material estimation, or physics-engine states.
\vspace{-4mm}
\paragraph{Limitations and Future Work.}
Our geometric fidelity is bounded by the V-JEPA encoder resolution and monocular depth accuracy. Sparse tokenization scales poorly to unbounded environments, moving cameras, or high-frequency deformations, and long rollouts accumulate noise; possibly mitigable via larger training chunks or feeding refined depth and V-JEPA latents back to the model. Our model also is not designed to generate unseen voxels. Future work includes constructing/generating full voxel volumes from image input and a universal V-JEPA decoder for photorealistic video.

\bibliography{reference}
\bibliographystyle{iclr2027/iclr2027_conference}

\clearpage
\appendix
\section{Implementation details} \label{app:implementation}

\subsection{Voxel Interpolation}\label{app:vox-interp}
Let $z_{t,u,v}^{(i)}\in\mathbb R^{1408}$ denote the
V-JEPA feature of patch $(u,v)$ in view $i$ at time $t$.
Using the estimated depth and camera parameters, we
unproject each patch to a 3D location $P_{t,u,v}^{(i)}$
and aggregate its feature into a fixed voxel grid. The grid bounds are determined from the initial observations and kept unchanged during prediction. The current implementation uses one camera per sample and expresses the grid in that camera's coordinate system.

Each point contributes to its eight neighbouring grid locations using trilinear weights. For voxel $j$, the aggregated feature is
\begin{equation}
 H_t(j)=
 \frac{\sum_{i,u,v}w_j(P_{t,u,v}^{(i)})\,z_{t,u,v}^{(i)}}
 {\max\!\left(\sum_{i,u,v}w_j(P_{t,u,v}^{(i)}),\delta\right)},
 \label{eq:trilinear-splat}
\end{equation}
where $w_j$ is the trilinear weight and $\delta$ is a small numerical constant. The sum runs over the views used for the sample; in our single-camera setup, there is only one view. Voxels with no contribution have zero features.

We store the full voxel state as $Z_t=[H_t,O_t,U_t]$, where $O_t$ denotes occupancy and $U_t$ denotes unobserved or occluded space. The prepared geometry targets are binary and mutually exclusive. Predicted geometry is continuous, with $O_t\geq0$ and $U_t\geq0$.

\subsection{Flow-matching objective}\label{app:flow-matching}
We model future voxel states using conditional flow matching. Each state $Z_t=[H_t,O_t,U_t]$ contains voxel features, occupancy, and unseen-space values. Given the clean future state $Z_{t+1}$, we sample Gaussian noise $\epsilon\sim\mathcal{N}(0,I)$ and an interpolation time $\tau\sim\mathcal{U}(0,1)$, and construct:
\begin{equation}
 Z_\tau=(1-\tau)Z_{t+1}+\tau\epsilon.
\end{equation}
Here $t$ indexes physical time, while $\tau$ controls the interpolation between the clean state and noise. The target velocity is:
\begin{equation}
 v^\star=\epsilon-Z_{t+1}.
\end{equation}
Conditioned on the available history, force inputs $u_{1:t+1}$, and physical time interval $\Delta t$, the predictor estimates:
\begin{equation}
 v_\theta=f_\theta\!\left(
 Z_{1:t},Z_\tau,u_{1:t+1},\Delta t,\tau
 \right).
\end{equation}
We train with the squared velocity error:
\begin{equation}
 \mathcal{L}_{\mathrm{FM}}
 =\mathbb{E}\!\left[
 \left\|v_\theta-(\epsilon-Z_{t+1})\right\|_2^2
 \right],
\end{equation}
where the expectation is over training sequences, noise, and interpolation times. The loss includes the feature, occupancy, and unseen channels, so all three are learned as continuous velocities.

At inference, we initialize the future state with Gaussian noise and integrate the predicted velocity from $\tau=1$ to $\tau=0$, keeping the conditioning inputs fixed. After integration, we constrain the geometry channels to $O\geq0$ and $U\geq0$. The generated state is then appended to the history to predict the next timestep.

\subsection{Latent projection}\label{app:render}
Given a generated voxel state $\hat Z_{t+1}=[\hat H_{t+1},\hat O_{t+1}, \hat U_{t+1}]$, we render a 2D latent map for camera $\Pi^{(i)}$. We use hard first-hit projection for final rendering and soft projection for the training loss. Each output location corresponds to a latent patch.

\paragraph{Ray-based depth projection.}
For final rendering, we cast one perspective ray per latent patch:
\begin{equation}
 \mathbf q_{u,v}(d)=\mathbf c+d\left(
 \frac{u-c_x}{f_x}\mathbf r+
 \frac{c_y-v}{f_y}\mathbf u+\mathbf f\right),
\end{equation}
where $\mathbf c$ is the camera centre, $\mathbf r,\mathbf u,\mathbf f$ are its right, up, and forward axes, and the intrinsics are scaled to latent-grid resolution. $d$ denotes camera-forward depth. We sample $N$ uniformly spaced depths between $d_{\mathrm{near}}$ and $d_{\mathrm{far}}$, the minimum and maximum forward depths of the bounding-box corners, with $d_{\mathrm{near}}$ clamped to zero. An empty interval produces no hits.

Let $j_k$ be the nearest voxel-grid location to sample $\mathbf q_{u,v}(d_k)$. We select the first in-bounds sample whose occupancy meets the threshold:
\begin{equation}
 k^\star=\min\{k:\mathbf q_{u,v}(d_k)\text{ is in bounds},\;
                  \hat O_{t+1}(j_k)\geq\theta_{\mathrm{occ}}\},
 \qquad
 \hat z_{t+1}^{(i)}(u,v)=\hat H_{t+1}(j_{k^\star}).
\end{equation}
The selected depth is $d_{k^\star}$. Rays with no hit have zero projected features and are marked invalid. This discrete visibility selection is used without gradient tracking.

\paragraph{Soft latent projection and projection loss}\label{app:softrast}
For training, we replace the hard visibility selection with front-to-back compositing. Projection supervision is applied to the estimated clean endpoint
\begin{equation}
 \bar Z_{t+1}=Z_\tau-\tau v_\theta.
\end{equation}

For each camera ray, we intersect the ray with the fixed bounding box and sample the intersecting interval at the midpoints of 32 equal segments. Rays that miss the volume have zero opacity and zero projected features.

For a given ray, let $h_k$ and $o_k$ be the features and occupancy obtained by trilinear interpolation at sample $k$, ordered from near to far. We convert occupancy to segment opacity as:
\begin{equation}
 \alpha_k=1-(1-o_k)^{\Delta s_k/\ell},
 \qquad
 T_k=\prod_{m<k}(1-\alpha_m),
 \qquad w_k=T_k\alpha_k,
\end{equation}
where $\Delta s_k$ is the segment length and $\ell$ is the minimum voxel spacing. Occupancy is capped slightly below one for numerical stability. This length correction avoids increasing opacity solely by subdividing a segment with constant occupancy. Out-of-bounds samples contribute zero opacity. The projected features and accumulated opacity are
\begin{equation}
 \hat z=\frac{\sum_k w_k H_k}{\max(\hat M,\delta)},
 \qquad \hat M=\sum_k w_k, 
\end{equation}
where $\delta$ is a small numerical constant. Normalizing the feature
weights avoids attenuating feature magnitudes on partially opaque rays; the opacity map itself is not normalized. The unseen channel does not contribute directly to opacity or feature compositing.

Let $h_p$ denote the reference feature encoded from the future RGB video and transformed into the same PCA space, where $p$ indexes the rendered patches. Let $M_p\in[0,1]$ be the fraction of valid depth observations in that patch and temporal tubelet, set to zero when the pooled 3D point lies outside the fixed volume. This is a depth-validity target, not an object segmentation mask. We use
\begin{equation}
 \mathcal L_{\mathrm{featproj}}=
 \frac{\sum_p M_p\|\hat z_p-z_p\|_2}
 {D\left(\sum_p M_p+\delta\right)},
 \qquad
 \mathcal L_{\mathrm{occproj}}=
 \frac{1}{|\mathcal P|}\sum_{p\in\mathcal P}(\hat M_p-M_p)^2,
\end{equation}
where $D$ is the feature dimension and $\mathcal P$ contains all rendered patches, including those with zero reference validity. For each predicted frame $r$, the loss combines the volumetric velocity losses with the two projection terms:
\begin{equation}
\mathcal L_r =
\mathcal L_{\mathrm{feat},r}
+\lambda_O\mathcal L_{O,r}
+\lambda_U\mathcal L_{U,r}
+\lambda_{\mathrm{fp}}\mathcal L_{\mathrm{featproj},r}
+\lambda_{\mathrm{op}}\mathcal L_{\mathrm{occproj},r}.
\label{eq:per-frame-loss}
\end{equation}
The full training objective averages $\mathcal L_r$ over the predicted frames and takes the expectation over training windows and sampled noise, as specified in Eq.~(4). We use $\lambda_{\mathrm{fp}}=\lambda_{\mathrm{op}}=1$.

\subsection{Model Architecture and training details}\label{app:model}
We use a spatial--temporal transformer with five blocks and a hidden dimension of 128. Each active voxel of the $25^3$ grid contains 1408 feature channels and two binary occupancy and unseen channels. We select occupied locations from the context and expand this set with two iterations of a $3\times3\times3$ neighbourhood dilation. A shared linear layer maps the resulting 1410-channel voxel vectors to 128-dimensional tokens. The same spatial locations are used across the context and noisy target frames.

We add spatial and temporal positional embeddings and an embedding of the force associated with each frame. Force and physical-time condition tokens are available to the spatial attention, while the noise level $\tau$ controls adaptive LayerNorm. Each block applies local spatial attention over a $5\times5\times5$ neighbourhood, a feed-forward network with hidden dimension 512 and GELU, and temporal self-attention at each voxel location. Attention uses eight heads, and the feed-forward hidden dimension is four times the token dimension. In our model, temporal attention includes the context and noisy target with a causal mask.

A small convolutional branch uses the latest history state concatenated with its difference from the preceding state.
A $1\times1\times1$ convolution first maps these inputs to hidden features at each voxel. Learned linear projections
of the force inputs, physical prediction interval, and noise levels of the two history states are added to these features and shared across spatial locations. A depthwise $3\times3\times3$ convolution then gathers local spatial information independently within each hidden channel. SiLU activations are applied before and after this spatial convolution, followed by a $1\times1\times1$ output projection to the joint state dimensions. We add this branch's output to the transformer's output to predict a joint velocity field for voxel features, occupancy, and unseen space.

Starting from two observed latent states, we train over 8 prediction steps. At each step, a state generated by integrating from fresh noise is appended to the history, with gradients retained through the rollout. The separate endpoint estimate obtained from a GT-interpolated noisy target is used only for supervision, including the soft projection losses; it is not used as generated history. 

We also corrupt history states at frame-dependent noise levels and provide those levels explicitly to the model. Active voxel regions are determined from the uncorrupted observed or generated history. At inference, initial observations remain clean, while generated history can receive the small corruption used by the stabilized sampling configuration.
We use AdamW with learning rate $3\times10^{-4}$, weight decay $10^{-4}$, and a ten-update linear warm-up followed by a constant learning rate. Batch size is one, with activation checkpointing and bfloat16 model computation; state updates and losses remain in float32. 

\subsection{Dataset and testing} \label{app:data}
We generated 1000 sequences of multiple simple rigid objects in a room bouncing and colliding with MuJoCo~\citep{todorov2012mujoco} of 3–6 rigid objects (spheres, cubes, and cylinders) interacting in a walled 8×5×3 m room, simulated at 30 fps with a 2 ms physics timestep. Each scene contains a 2.5 s active rollout (38 latent frames at 30 fps), during which a randomized 3D impulse force is applied to either a single target object (local scope, 50 \%) or all objects simultaneously (global scope, 50 \%). Objects vary in size, mass, color, and initial airborne state (1–3 objects are in flight at the start). Each sequence is rendered from 5 fixed cameras at 480×480 resolution. For the PhysInOne~\citep{zhou2026physinonevisualphysicslearning} dataset, we only found the validation dataset with 200 videos from the publicly available PhysInOne repository, each rendered from 12 stationary cameras at 1120×1120 and 30 fps with per-frame ground-truth depth. Each trajectory spans about 150 frames (5 s) and involves two or three simultaneously active physics phenomena drawn from a large vocabulary, including rigid-body collisions, liquid dynamics, and wind-driven motion. The PhysGaia~\citep{kim2026physgaia} benchmark covers four continuum-material categories in 17 scenes: smoke/gas, fluid/liquid, cloth/textile, and viscoelastic solids. Each scene is rendered from 4 views at 720×960 and 24 fps; sequences are 240 frames (10s). No ground-truth depth is available, so depth is estimated with MoGe~\citep{wang2025moge}. For all datasets, we used 5\% of the data for testing, and specifically hand-picked the ones with fluid and smoke interactions.

We acknowledge that evaluating in V-JEPA space may favor methods whose representations are closer to V-JEPA-style features. However, we argue that latent-space evaluation provides a meaningful complementary criterion for video generation and prediction. A physically plausible generated video should not only match the target at the pixel level or satisfy high-level VLM judgments, but should also remain close to the target in a temporally structured video representation space. We therefore measure prediction error in V-JEPA feature space. V-JEPA is a natural choice because its embeddings are learned through predictive video objectives and are temporally aligned, allowing the metric to compare dynamic scene evolution rather than isolated frame appearance.

\section{Additional experiments}
\subsection{Latent prediction}\label{app:exp-latent}

Generally, the multi-camera errors in \Cref{tab:latent_protocols} are higher than the single-camera ones because more angles are introduced to calculate loss, while the model and baselines' input is mostly single-camera. Leaving out multi-camera PhysGaussian, our model shows a comparatively small increase when moving from single-camera to multi-camera evaluation, indicating some cross-view consistency despite using single-camera information during training. CogVideoX and PhysGen are competitive in some single-camera settings, especially on simpler rigid-body scenes, e.g., CLEVRER. But both CogVideoX and PhysGen degrade when moving to multi-camera evaluation. This shows that although CogVideoX is trained to reduce implicit embedding loss, and PhysGen follows physical laws, they are 2D methods and would have a large drop in performance for 3D consistency. Ours-GT performed the best with the overall loss, but since the dynamics is trained on GT, the 3D feature loss is consistently higher than Ours-estimate. 

\begin{table*}[t]
\caption{Latent prediction results under different camera-depth evaluation protocols. Each entry reports \textbf{2D / 3D latent L2 loss}~(\(\downarrow\)). Columns specify the complete evaluation condition: single-camera or multi-camera prediction with either ground-truth or estimated depth. \textsc{N/A} indicates that the dataset does not provide ground-truth depth, when available. The GT depth columns for CLEVRER are tested with our synthetic dataset, see Appendix~\ref{app:implementation}.
}
\vspace{2mm}
\centering

\tiny

\begin{adjustbox}{width=1.0\textwidth,center}
\setlength{\tabcolsep}{4pt}
\renewcommand{\arraystretch}{1.08}
\begin{tabular}{clcccc}
\toprule
\multirow{2}{*}{Data} 
& \multirow{2}{*}{Method}
& \multicolumn{2}{c}{Single-camera}
& \multicolumn{2}{c}{Multi-camera} \\
\cmidrule(lr){3-4} \cmidrule(lr){5-6}
& 
& GT depth & estimated depth
& GT depth & estimated depth \\
\midrule
\multirow{7}{*}{\rotatebox{90}{\makecell{
            CLEVRER\\
            \citep{yi2019clevrer}
        }}}
& CogVideoX     & 2.98 $\pm$ 0.68 / 0.22 $\pm$ 0.12 & 2.98 $\pm$ 0.68 / 0.56 $\pm$ 0.13 & 3.87 $\pm$ 0.58 / 1.50 $\pm$ 0.11 & 4.14 $\pm$ 0.49 / 1.18 $\pm$ 0.07\\
& 
PhysGen        
& 1.91 $\pm$ 0.08 / 0.17 $\pm$ 0.02 & 2.97 $\pm$ 0.19 / 0.56 $\pm$ 0.03 & 3.57 $\pm$ 0.17 / 1.89 $\pm$ 0.11 & 4.12 $\pm$ 0.49 / 1.24 $\pm$ 0.19\\
& PhysGaussian   & 4.13 $\pm$ 0.08 / 0.56 $\pm$ 0.01 & 4.39 $\pm$ 0.08 / 0.55 $\pm$ 0.03 & 4.34 $\pm$ 0.09 / 1.18 $\pm$ 0.07 & 4.21 $\pm$ 0.09 / 1.14 $\pm$ 0.08 \\
& PhysCtrl       & 2.96 $\pm$ 0.37 / 0.35 $\pm$ 0.05 & \secondbest{2.84 \pm 0.23} / 0.57 $\pm$ 0.03 & 3.32 $\pm$ 0.31 / 1.01 $\pm$ 0.07 & 3.21 $\pm$ 0.20 / 1.15 $\pm$ 0.08 \\
& 2D baseline  & 1.39 $\pm$ 0.28 / 0.32 $\pm$ 0.08 & 1.53 $\pm$ 0.71 / 0.42 $\pm$ 0.12 & 3.19 $\pm$ 0.37  / 1.20 $\pm$ 0.27 & 3.69 $\pm$ 0.59 / 1.10 $\pm$ 0.11 \\
& Ours-GT        & \best{0.98 \pm 0.07} / \best{0.02 \pm 0.00} & \best{1.01 \pm 0.10} / \secondbest{0.39 \pm 0.01} & \best{1.12 \pm 0.09} / \best{0.82 \pm 0.05} & \secondbest{2.51 \pm 0.11} / \secondbest{0.91 \pm 0.05} \\
& Ours-estimate  & $\secondbest{1.26 \pm 0.13} / \secondbest{0.25 \pm 0.03}$ & $\best{1.01 \pm 0.23} / \best{0.36 \pm 0.04}$ & $\secondbest{2.50 \pm 0.15 } / \secondbest{0.99 \pm 0.09}$ & $\best{2.25 \pm 0.09} / \best{0.90 \pm 0.09}$ \\ 
\midrule

\multirow{7}{*}{\rotatebox{90}{\makecell{PhysInOne\\\citep{zhou2026physinonevisualphysicslearning}}}}
& CogVideoX      & \secondbest{3.02 \pm 0.32} / \secondbest{0.53 \pm 0.16} & \secondbest{3.08 \pm 0.33} / 0.75 $\pm$ 0.20 & 4.05 $\pm$ 0.31 / 1.01 $\pm$ 0.11 & 4.00 $\pm$ 0.39 / 1.00 $\pm$ 0.19 \\
& PhysGen        & 3.47 $\pm$ 0.59 / 0.52 $\pm$ 0.15 & 3.61 $\pm$ 0.27 / 0.71 $\pm$ 0.19 & 3.98 $\pm$ 0.37 / 0.87 $\pm$ 0.20 & 4.12 $\pm$ 0.91 / 1.02 $\pm$ 0.49 \\
& PhysGaussian   & 3.98 $\pm$ 0.22 / 1.02 $\pm$ 0.22 & 3.95 $\pm$ 0.89 / 0.89 $\pm$ 0.20 & 4.57 $\pm$ 0.21 / 0.90 $\pm$ 0.15 & 4.62 $\pm$ 0.23 / 1.04 $\pm$ 0.21 \\
& PhysCtrl       & 4.24 $\pm$ 0.37 / 0.66 $\pm$ 0.31 & 4.23 $\pm$ 0.51 / 0.81 $\pm$ 0.21 & 4.57 $\pm$ 0.35 / 0.84 $\pm$ 0.17 & 4.34 $\pm$ 0.38 / 0.98 $\pm$ 0.21 \\
& 2D baseline & 2.99 $\pm$ 0.13 / 0.55 $\pm$ 0.19 & 2.25 $\pm$ 0.23 / 0.82 $\pm$ 0.19 & 3.98 $\pm$ 0.11 / 0.99 $\pm$ 0.12 & 4.14 $\pm$ 0.38 / 1.00 $\pm$ 0.50 \\
& Ours-GT        & \best{0.43 \pm 0.05} / \best{0.06 \pm 0.01} & \secondbest{3.08 \pm 0.17} / \secondbest{0.63 \pm 0.14} & \best{2.62 \pm 0.10} / \best{0.49 \pm 0.07} & \secondbest{2.62 \pm 0.17} / \secondbest{0.51 \pm 0.14} \\
& Ours-estimate  & \secondbest{3.02 \pm 0.17} / \secondbest{0.53 \pm 0.14} & \best{1.67 \pm 0.13} / \best{0.42 \pm 0.14} & \secondbest{3.51 \pm 0.17} / \secondbest{0.66 \pm 0.14} & \best{2.59 \pm 0.21} / \best{0.49 \pm 0.11} \\
\midrule

\multirow{7}{*}{\rotatebox{90}{\makecell{
        PhysGaia\\ \citep{kim2026physgaia}}}}
& CogVideoX      & \textsc{N/A} & 3.71 $\pm$ 0.72 / 0.96 $\pm$ 0.32 & \textsc{N/A} & 3.72 $\pm$ 0.72 / 1.57 $\pm$ 0.47 \\
& PhysGen        & \textsc{N/A} & 4.89 $\pm$ 0.50 / 1.15 $\pm$ 0.32 & \textsc{N/A} & 5.23 $\pm$ 0.52 / 1.68 $\pm$ 0.54 \\
& PhysGaussian   & \textsc{N/A} & 3.98 $\pm$ 1.16 / 0.96 $\pm$ 0.33 & \textsc{N/A} & 3.95 $\pm$ 1.21 / 1.58 $\pm$ 0.48 \\
& PhysCtrl       & \textsc{N/A} & 4.25 $\pm$ 1.05 / 0.97 $\pm$ 0.32 & \textsc{N/A} & 4.25 $\pm$ 1.09 / 1.58 $\pm$ 0.48 \\
& 2D baseline  & \textsc{N/A} & 2.16 $\pm$ 0.92 / 0.95 $\pm$ 0.61 & \textsc{N/A} & 4.97 $\pm$ 0.99 / 1.47 $\pm$ 0.22 \\
& Ours-GT        & \textsc{N/A} & \secondbest{1.84 \pm 0.78} / \secondbest{0.47 \pm 0.21} & \textsc{N/A} & \secondbest{2.10 \pm 0.50} / \secondbest{1.15 \pm 0.35} \\
& Ours-estimate  & \textsc{N/A} & \best{1.66 \pm 0.50} / \best{0.41 \pm 0.21} & \textsc{N/A} & \best{1.99 \pm 0.51} / \best{1.00 \pm 0.09} \\
\bottomrule
\end{tabular}
\label{tab:latent_protocols}
\end{adjustbox}
\end{table*}

\subsection{Flow estimation}\label{app:flow}
%%%%%% old latent flow extraction
\paragraph{Experiment settings.}
We evaluate the predicted video latents against ground-truth trajectories by comparing their latent flow consistency. We compare both the 2D and 3D flow predictions. This tests dynamics by evaluating predicted video latents against ground-truth trajectories using latent flow consistency. For baseline methods, each video frame generated is encoded into feature space with the VJEPA encoder and projected into 3D space for 3D evaluation. For our models, we directly generate the latent according to past frames, and projected to 2D for 2D evaluations. Given two consecutive latent frames, we estimate a per-patch displacement field by normalizing and comparing the cosine distance within a local window in the next frame. For each trajectory, we compute consecutive flow pairs from both the predicted and ground-truth latent sequences, then measure the total error accumulated between them by calculating their displacement differences. A lower error indicates that the predicted video exhibits motion patterns, in 2D or 3D, more consistent with the ground truth. The updated flow evaluation is reported on CLEVRER, PhysGaia, PhysInOne, and the 30 zero-shot samples from the unseen Kubric MOVI-C dataset. For these experiments, we report Ours-Estimate, our longer-horizon model trained with diffusion forcing and history-guided diffusion. For 3D flow loss, we omit the 2D baseline as it requires an extra z channel that cannot be supplied.

\begin{table*}[h!]
\centering
\caption{
2D latent-flow estimation error across the evaluation datasets.
Lower values indicate greater consistency between the predicted and
ground-truth 2D motion. Kubric MOVI-C is evaluated zero-shot.
Best results are shown in bold and second-best results are underlined.
}
\label{tab:flow_2d_comparison}

\small
\begin{tabular}{lcccc}
\toprule
Method
& CLEVRER $\downarrow$
& PhysGaia $\downarrow$
& PhysInOne $\downarrow$
& Kubric MOVI-C $\downarrow$ \\
\midrule

CogVideoX
& $0.492 \pm 0.031$
& $0.824 \pm 0.100$
& $0.753 \pm 0.065$
& $0.667 \pm 0.094$ \\

PhysGen
& $\underline{0.315 \pm 0.022}$
& $0.737 \pm 0.105$
& $0.637 \pm 0.110$
& $0.738 \pm 0.133$ \\

PhysGaussian
& $0.407 \pm 0.007$
& $0.682 \pm 0.134$
& $\underline{0.405 \pm 0.083}$
& $0.836 \pm 0.126$ \\

PhysCtrl
& $0.651 \pm 0.017$
& $0.778 \pm 0.127$
& $0.694 \pm 0.084$
& $0.633 \pm 0.091$ \\

2D baseline
& $0.383 \pm 0.020$
& $\underline{0.534 \pm 0.160}$
& $0.435 \pm 0.055$
& $\underline{0.563 \pm 0.109}$ \\

Ours-Estimate
& $\mathbf{0.273 \pm 0.009}$
& $\mathbf{0.464 \pm 0.118}$
& $\mathbf{0.373 \pm 0.031}$
& $\mathbf{0.488 \pm 0.151}$ \\

\bottomrule
\end{tabular}

\end{table*}

\begin{figure}[h]
    \centering
    \includegraphics[width=0.32\linewidth]{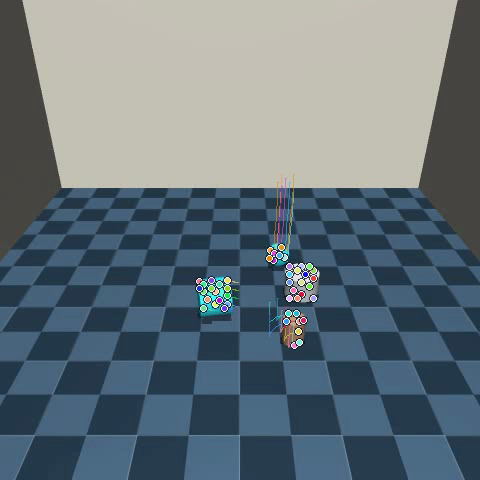}\hfill
    \includegraphics[width=0.32\linewidth]{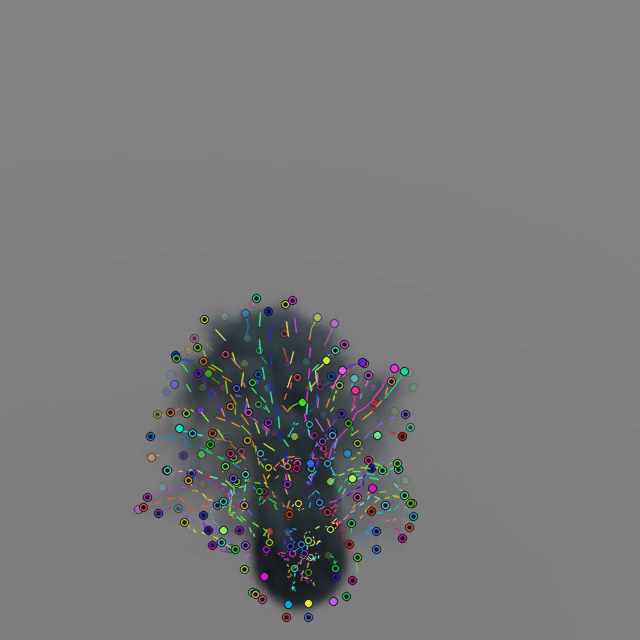}\hfill
    \includegraphics[width=0.32\linewidth]{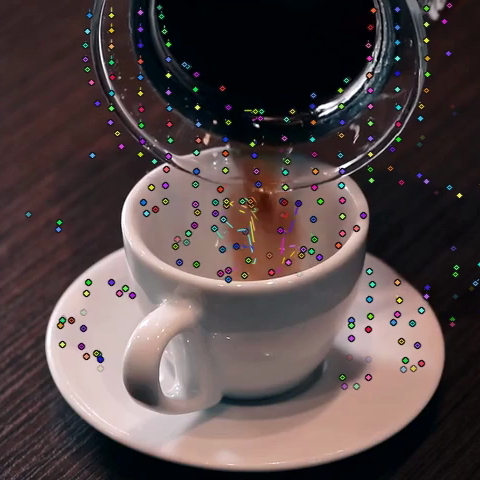}

    \par\medskip 

    \includegraphics[width=0.32\linewidth]{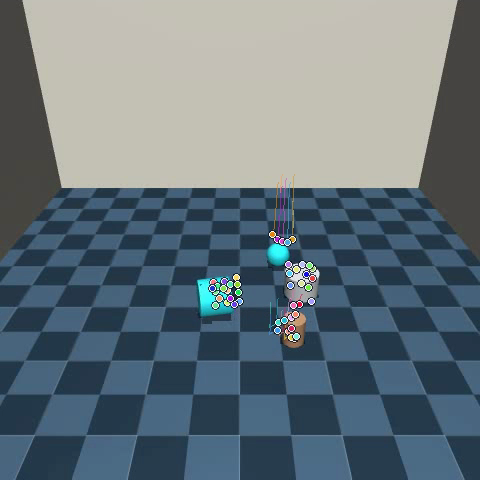}\hfill
    \includegraphics[width=0.32\linewidth]{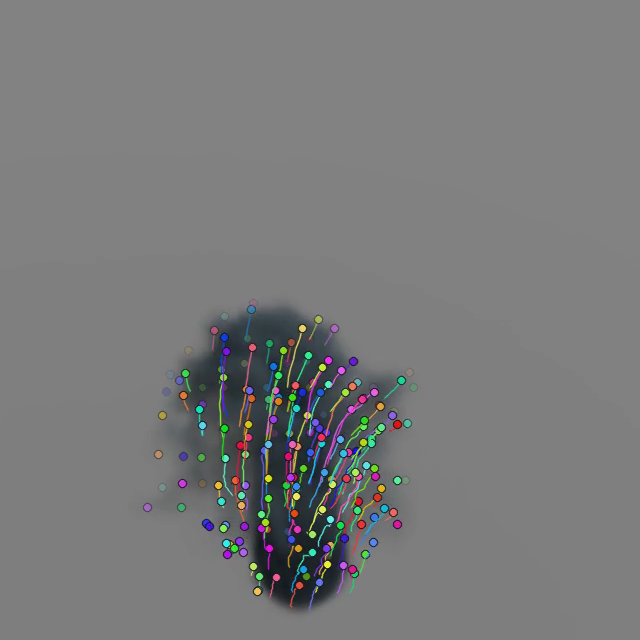}\hfill
    \includegraphics[width=0.32\linewidth]{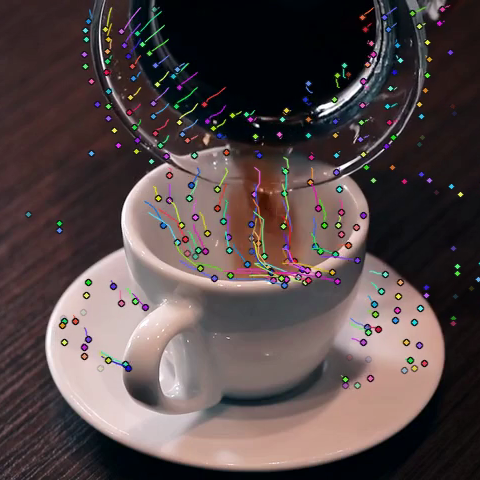}

    \caption{Qualitative comparison of 2D flow extraction in \Cref{sec:exp-flow}.
        Ground-truth flow is shown on the top, and our estimated flow on the bottom, with flow arrows overlaid on the V-JEPA features for visualization.
        Our voxel dynamics model is not trained with an additional 2D flow loss, so this comparison illustrates how the learned representation can be repurposed for motion estimation. See supplemental webpage for videos. 
        }
    \label{fig:flow_2d_visualization}
\end{figure}

\paragraph{Results.}
We report the mean 2D and 3D flow errors across the three evaluation datasets and the zero-shot Kubric MOVI-C dataset in \Cref{tab:flow_2d_comparison,tab:flow_3d_estimation}. Ours-Estimate obtains the lowest 2D and 3D flow error across all four datasets, including the unseen Kubric MOVI-C samples. The 2D baseline remains competitive in several 2D evaluations but is consistently outperformed by our 3D model, supporting the contribution of the 3D projection beyond optimizing the same latent-space objective. The zero-shot Kubric result further evaluates whether the learned dynamics transfer to scenes outside the training datasets. We also show a visualization of the trajectory from the optical flow head in \Cref{sec:exp-flow} (not the per-patch displacement field) in \Cref{fig:flow_2d_visualization}.

\begin{table*}[t!]
\centering
\caption{3D latent-flow estimation error across the evaluation datasets. Lower values indicate greater consistency between the predicted and ground-truth 3D motion. Kubric MOVI-C is evaluated zero-shot. Best results are shown in bold and second-best results are underlined.}
\label{tab:flow_3d_estimation}
\scriptsize
\setlength{\tabcolsep}{7pt}
\begin{tabular}{lcccc}
    \toprule
    Method
    & CLEVRER $\downarrow$
    & PhysGaia $\downarrow$
    & PhysInOne $\downarrow$
    & Kubric MOVI-C $\downarrow$ \\
    \midrule

    CogVideoX
    & $0.6188 \pm 0.0682$
    & $0.7011 \pm 0.0190$
    & $0.7627 \pm 0.0679$
    & $0.7353 \pm 0.0253$ \\

    PhysGen
    & $0.6827 \pm 0.0250$
    & $0.7007 \pm 0.0309$
    & $0.7480 \pm 0.0700$
    & $0.7214 \pm 0.0215$ \\

    PhysGaussian
    & $0.7277 \pm 0.0061$
    & $0.7227 \pm 0.0612$
    & $\underline{0.7064 \pm 0.0996}$
    & $0.7900 \pm 0.0722$ \\

    PhysCtrl
    & $\underline{0.5335 \pm 0.0966}$
    & $\underline{0.6325 \pm 0.0913}$
    & $0.7113 \pm 0.1048$
    & $\underline{0.6909 \pm 0.0991}$ \\

    Ours-Estimate
    & $\mathbf{0.3564 \pm 0.0146}$
    & $\mathbf{0.5640 \pm 0.0445}$
    & $\mathbf{0.5370 \pm 0.0313}$
    & $\mathbf{0.5194 \pm 0.0520}$ \\

    \bottomrule
\end{tabular}
\end{table*}

\subsection{Occupancy evaluation}

\paragraph{Experiment settings.}
In addition to flow, we directly evaluate the predicted 3D occupancy against the ground-truth occupied voxels. We report occupancy IoU as the primary geometric measure and the ratio between predicted and ground-truth occupied volume as an additional diagnostic. Higher IoU indicates better spatial agreement. A mass ratio closer to one is informative primarily for approximately closed, mass-conserving objects; it need not remain constant for phenomena such as smoke. Moreover, a mass ratio close to one does not imply that the occupied volume is placed correctly, so we interpret it jointly with occupancy IoU. We evaluate Ours-Estimate using occupancy thresholds of $0.4$, $0.5$, and $0.6$. The same evaluation is performed on CLEVRER, PhysGaia, PhysInOne, and the zero-shot Kubric MOVI-C samples. For the baseline video generators, the occupancy is extracted from the 3D voxels that are unprojected using estimated depth.

\begin{table*}[h!]
\centering
\caption{
Occupancy-based evaluation. Each entry reports occupancy IoU and mass ratio.
Higher IoU is better, while a mass ratio closer to one is better.
Our model is evaluated using three different occupancy thresholds.
Best results are shown in bold and second-best results are underlined.
}
\label{tab:occupancy_evaluation}

\fontsize{7.5pt}{9pt}\selectfont
\setlength{\tabcolsep}{7pt}
\renewcommand{\arraystretch}{1.15}

\begin{tabular}{lcc}
\toprule
Method
& CLEVRER: IoU $\uparrow$ / mass ratio $\sim 1$
& PhysGaia: IoU $\uparrow$ / mass ratio $\sim 1$ \\
\midrule

CogVideoX
& $0.4168 \pm 0.0962$ / $1.1006 \pm 0.0806$
& $0.0846 \pm 0.0536$ / $\mathbf{0.7661 \pm 0.4417}$ \\

PhysGen
& $0.5341 \pm 0.1057$ / $\mathbf{1.0165 \pm 0.0205}$
& $0.1197 \pm 0.0767$ / $1.3260 \pm 0.8176$ \\

PhysGaussian
& $0.3450 \pm 0.0058$ / $0.8298 \pm 0.0175$
& $0.1076 \pm 0.0903$ / $0.6329 \pm 0.7830$ \\

PhysCtrl
& $0.1478 \pm 0.0197$ / $0.6849 \pm 0.0163$
& $0.1129 \pm 0.0489$ / $\underline{0.6874 \pm 0.4790}$ \\

Ours-Estimate (0.4 threshold)
& $\underline{0.9119 \pm 0.0384}$ / $\underline{1.0789 \pm 0.0361}$
& $0.7059 \pm 0.0797$ / $1.3834 \pm 0.1076$ \\

Ours-Estimate (0.5 threshold)
& $\underline{0.9119 \pm 0.0384}$ / $\underline{1.0789 \pm 0.0361}$
& $\underline{0.7061 \pm 0.0798}$ / $1.3827 \pm 0.1076$ \\

Ours-Estimate (0.6 threshold)
& $\mathbf{0.9120 \pm 0.0384}$ / $\underline{1.0789 \pm 0.0361}$
& $\mathbf{0.7064 \pm 0.0798}$ / $1.3821 \pm 0.1075$ \\

\bottomrule
\end{tabular}

\vspace{0.25cm}

\begin{tabular}{lcc}
\toprule
Method
& PhysInOne: IoU $\uparrow$ / mass ratio $\sim 1$
& Kubric MOVI-C: IoU $\uparrow$ / mass ratio $\sim 1$ \\
\midrule

CogVideoX
& $0.0880 \pm 0.0183$ / $0.2612 \pm 0.0403$
& $0.0835 \pm 0.0634$ / $0.5623 \pm 0.3523$ \\

PhysGen
& $0.2155 \pm 0.0212$ / $\underline{1.2152 \pm 0.5123}$
& $0.4243 \pm 0.0422$ / $\mathbf{1.0233 \pm 0.9457}$ \\

PhysGaussian
& $0.1151 \pm 0.0147$ / $\mathbf{0.7853 \pm 0.2199}$
& $0.2043 \pm 0.0323$ / $0.5352 \pm 0.2334$ \\

PhysCtrl
& $0.1000 \pm 0.0115$ / $0.7008 \pm 0.0458$
& $0.1290 \pm 0.0720$ / $\underline{0.6982 \pm 0.9228}$ \\

Ours-Estimate (DF), $0.4$
& $0.7538 \pm 0.0409$ / $1.3089 \pm 0.0643$
& $0.5286 \pm 0.0454$ / $1.6276 \pm 0.6049$ \\

Ours-Estimate (DF), $0.5$
& $\underline{0.7541 \pm 0.0408}$ / $1.3101 \pm 0.0646$
& $\underline{0.5287 \pm 0.0454}$ / $1.6119 \pm 0.5947$ \\

Ours-Estimate (DF), $0.6$
& $\mathbf{0.7543 \pm 0.0408}$ / $1.3095 \pm 0.0644$
& $\mathbf{0.5288 \pm 0.0454}$ / $1.5951 \pm 0.5843$ \\

\bottomrule
\end{tabular}

\end{table*}

\paragraph{Results.}
Ours-Estimate obtains substantially higher occupancy IoU than the baselines across all four datasets, including the unseen Kubric MOVI-C samples. The mass-ratio results should not be treated as a standalone ranking: a ratio close to one may coincide with incorrect spatial occupancy, while non-conservative phenomena such as smoke need not preserve total occupied volume. We therefore use occupancy IoU as the primary geometric measure and interpret mass ratio jointly with it.
\subsection{RGB evaluations}
\subsubsection{Physics IQ benchmark}\label{app:physicsiq-full}
\begin{table}[h]
    \centering
    \caption{Physics IQ scores in four criteria.}
    \label{tab:physicsIQ_full}
    \scriptsize
    \begin{tabular}{lccccc}
        \toprule
        Method & Spatiotemporal IoU $\uparrow$ & Spatial IoU $\uparrow$ & Weighted Spatial IoU $\uparrow$ & MSE $\downarrow$ & score $\uparrow$\\
        \midrule
         CogVideoX & 0.1782 & 0.3835 & 0.2273 & 0.0259 & 0.237 \\
         PhysGen & 0.2110 & 0.2212 & 0.1199 & 0.0210 & 0.163 \\
         PhysGaussian & \best{0.2894} & 0.3599 & 0.2161 & \secondbest{0.0162} & \secondbest{0.272} \\
         PhysCtrl & \secondbest{0.2725} & 0.3939 & 0.2504 & 0.0248 & 0.281 \\
         Wan-move (no tracks) & 0.1373 & \best{0.4648} & \secondbest{0.2521} & 0.0256 & 0.259 \\
         Ours-Estimate & 0.2643 & \best{0.5566} & \best{0.3752} & \best{0.0152} & \best{0.384} \\
    \end{tabular}
\end{table}

\subsubsection{Signal-to-noise ratio}

In order to get back to pixels, we decode predicted latents to RGB in two ways:
(i) a scene-specific lightweight V-JEPA-to-video head trained on the CLEVRER-style data, and (ii) a
trajectory flow-conditioned video~\citep{chu2025wan} generator driven by our predicted flow. We
report PSNR against reference videos (\Cref{tab:rgb_psnr}); see supplemental for the qualitative videos. We omit FVD, as the number of decoded clips is too
small for a meaningful distribution-level estimate.

\begin{table}[h]
\centering
\caption{RGB reconstruction fidelity PSNR (dB) of decoded predictions.}
\label{tab:rgb_psnr}
\scriptsize
\begin{tabular}{lcccc}
\toprule
Method & CLEVRER $\uparrow$ & PhysInOne $\uparrow$ & PhysGaia $\uparrow$ & Kubric $\uparrow$  \\
\midrule
CogVideoX      & $16.40 \pm 0.53$ & \best{20.46 \pm 4.07} & $16.96 \pm 7.11$ & $17.19 \pm 3.00$ \\
PhysGen        & $24.71 \pm 2.30$ & $19.17 \pm 3.39$ & $16.45 \pm 8.71$ & $16.36 \pm 1.48$ \\
PhysGaussian   & $25.87 \pm 1.64$ & $\secondbest{19.92 \pm 3.52}$ & $15.99 \pm 5.94$ & $17.87 \pm 3.11$ \\
PhysCtrl       & $18.21 \pm 5.14$ & $19.63 \pm 2.19$ & $16.80 \pm 5.95$ & $17.12 \pm 3.04$ \\
2D baseline    & $26.42 \pm 1.48$ & $17.20 \pm 1.36$ & $17.02 \pm 6.27$ & $17.30 \pm 1.67$ \\
Ours-Estimate (decoder)  & \best{29.83 \pm 2.38} & $18.59 \pm 1.34$ & \best{18.39 \pm 6.78} & \best{17.89 \pm 2.54}\\
Ours-Estimate (wan-move)  & $22.85 \pm 2.62$ & $14.14 \pm 4.30$ & $\secondbest{18.57 \pm 3.83}$ & $15.00 \pm 2.00$\\
\bottomrule
\end{tabular}
\end{table}

\paragraph{Results.}
Decoded predictions preserve object identity and motion, giving the best PSNR among
compared methods by 5 dB. The video-from-flow path produces sharper textures at
comparable motion accuracy, showing the predicted dynamics are precise enough to
drive a larger high-fidelity generator.

\subsection{Additional ablations}\label{app:ablations}
\paragraph{Effect of depth or geometry estimators.}
In the development of our pipeline, we have tested different depth estimators for videos or multiple frames, and we show the results in \Cref{tab:ablation_estimators}. We have tested a static scene reconstruction model, VGGT~\citep{wangVGGTVisualGeometry2025a}, the combination of depth estimator~\citep{video_depth_anything} and camera estimator~\citep{mast3r}, and a depth and camera estimator, MoGe~\citep{wang2025moge}. The choice of estimator has a substantial impact on all metrics, showing that the lifting process is highly sensitive to the quality of the underlying geometric estimates. Once 2D features are lifted into 3D, geometric errors directly translate into incorrect voxel occupancy and degraded feature alignment. We found that MoGe performs best across all metrics, and thus used it for our main results.

\begin{table*}[h]
\centering
\caption{Effect of the depth estimator choice.}
\setlength{\tabcolsep}{5pt}
\begin{tabular}{lccc}
\toprule
\textbf{Method} & \textbf{IoU \%} $\uparrow$ &  \textbf{2D feat} $\downarrow$ & \textbf{3D feat} $\downarrow$ \\
\midrule
VidDepA;Mast3r & 38.64 & 3.19 & 1.76 \\
VGGT & 61.03 & 2.19 & 0.88 \\
MoGe & 87.13 & 1.78 & 0.61 \\
\bottomrule
\end{tabular}
\label{tab:ablation_estimators}
\end{table*}

\paragraph{Number of views.}
We test the effect of view coverage by varying both the number of input views in \Cref{tab:ablation_views}. We test our model on different numbers of input views at inference time, then compute the IoU and feature loss for single-camera and multiple-camera settings. In the multi-camera evaluation setting, increasing the view coverage has a limited effect on IoU. We suspect that this is because, although the voxel construction gains higher quality for multiple camera angles as input (suggested in the single camera column), there are more entries for noise and loss aggregates; thus, the effects cancel out. This still supports the use of diverse camera viewpoints when constructing the voxel state, as they reduce occlusion ambiguity and encourage a more complete scene representation.

\begin{table*}[h]
\centering

\caption{Effect of the number of views and view-angle span on 2D/3D feature loss.}
\centering
\setlength{\tabcolsep}{4pt} 
\begin{tabular}{lcccc}
\toprule
\textbf{views} 
& \textbf{Angle} 
& \multicolumn{3}{c}{\textbf{multi-camera / single-camera}} \\
\cmidrule(lr){3-5}
& 
& \textbf{IoU \% } $\uparrow$ 
& \textbf{2D feat} $\downarrow$ 
& \textbf{3D feat} $\downarrow$ \\
\midrule
1 & $0^\circ$ & 71.87 / 80.00 & 2.99 / 2.23 & 1.47 / 0.61  \\
2 & $<45^\circ$ & 71.07 / 87.62 & 2.82 / 2.22 & 1.48 / 0.53  \\
max & $>90^\circ$ & 71.50 / 90.25 & 2.18 / 1.43 & 1.52 / 0.32  \\
\bottomrule
\end{tabular}
\label{tab:ablation_views}
\end{table*}

\paragraph{Quality vs computation for voxel grid size.}
We studied the effect of voxel grid resolution on reconstruction quality and training memory in \Cref{tab:ablation_grid_size}. We show the 2D and 3D feature loss, in addition to the occupancy IoU. As expected, increasing the grid resolution consistently improves all quality metrics, indicating that the voxel representation benefits from finer spatial discretizations. But higher voxel resolution also introduces a clear memory-quality trade-off. We speculate that this gain in quality is due to the gaps between the V-JEPA latent patch size 30 and the grid size. Unfortunately, higher grid sizes cannot be tested due to limited memory.

\begin{table*}[h]
\caption{Trade-off between quality and computation under different voxel grid resolutions.}
\centering
\footnotesize
\setlength{\tabcolsep}{5pt}
\begin{tabular}{ccccc}
\toprule
\textbf{Voxel Grid Size} & \textbf{Train Mem.} $\downarrow$ & \textbf{IoU} $\uparrow$ & \textbf{2D feature Loss} $\downarrow$ & \textbf{3D feature Loss} $\downarrow$  \\
\midrule
$10^3$ & 1.44 GB & 51.74\%  & 2.75 & 1.60\\
$15^3$ & 4.70 GB & 65.14\%  & 2.63 & 1.55\\
$20^3$ & 11.06 GB & 69.79\%  & 2.02 & 0.99\\
$25^3$ & 23.45 GB & 87.13\% & 1.78 & 0.61 \\
\bottomrule
\end{tabular}
\label{tab:ablation_grid_size}
\end{table*}

\paragraph{Projection loss.}
We compare the model's performance \textit{with} and \textit{without} training with our projection loss in \Cref{tab:ablation_projection_loss}. The results show that the projection loss is a key component of the proposed representation learning objective. Although it is computationally expensive to add, it corresponds to a large improvement in both geometric overlap and feature consistency. Without this constraint, the model obtains a weaker voxel representation, suggesting that volumetric supervision from voxels from monocular views alone is insufficient for reliable feature reconstruction. 

\begin{table*}[h]
\centering
\caption{Effect of the projection loss on reconstruction quality and memory usage.}
\setlength{\tabcolsep}{4pt}
\begin{tabular}{lcccc}
\toprule
\textbf{Proj.} & \textbf{Mem.} $\downarrow$ & \textbf{IoU\%} $\uparrow$ & \textbf{2D feat} $\downarrow$ & \textbf{3D feat} $\downarrow$ \\
\midrule
w/o & 100\% & 69.47 & 3.24 & 1.75   \\
w/ & 110\% & 87.13 & 1.78 & 0.61  \\
\bottomrule
\end{tabular}
\label{tab:ablation_projection_loss}

\end{table*}

\paragraph{Additional voxel channels.}
We also tested the importance of our occupancy channels in \Cref{tab:ablation_voxel_channels}, under the monocular setting. The feature-only variant performs poorly, with a 2D feature loss of $4.97$ and a 3D feature loss of $2.42$. This is intuitive because without a depth/occupancy indicator, the whole voxel grid would be filled with active voxels; thus, the 3D features cannot learn accurate geometry information, and the projection would be blocked by the voxels nearest to the camera. Adding an \texttt{occupancy} channel substantially improves the representation. The further added \texttt{observed} channel indicates that parts of the unoccupied voxels might not be empty, but unseen. This further improves our model's performance considerably.

\begin{table*}[h]
\centering
\small
\caption{Effect of choice of channels.}
\vspace{.025in}
\setlength{\tabcolsep}{5pt}
\begin{tabular}{lcccc}
\toprule
\textbf{Voxel State} & \textbf{\# ch.} & \textbf{IoU\%} $\uparrow$ &  \textbf{2D feat} $\downarrow$ & \textbf{3D feat} $\downarrow$ \\
\midrule
Features only & 1408 & \textsc{N/A} & 4.97 & 2.42  \\
Features with occupancy channels & 1409 & $67.52$ & 2.96 & 1.02 \\
Features, occupancy and observed channels & 1410 & $87.13$ & 1.78 & 0.61 \\
\bottomrule
\end{tabular}
\label{tab:ablation_voxel_channels}

\end{table*}

\paragraph{Occupancy and occlusion losses.}\label{app:occupancy}
We ablate the occupancy and occlusion losses to measure their contribution to spatial consistency and motion prediction. All variants in this comparison are trained for 5,000 steps, and we report occupancy IoU together with 2D and 3D flow error. 
Removing either loss degrades the full model, while removing both produces the largest 2D-flow error. These results support the contribution of both terms to the predicted occupancy and motion.

\paragraph{Is a generative model necessary?}\label{app:regressVsGenerative}

In \Cref{tab:regression_flow_matching} we evaluate the importance of a generative model. Because of partial observation and lack of exact material parameters, our prediction task is indeed reasoning over a distribution. Hence, the generative model performs better. 

\begin{table}[h!]
\centering
\caption{Deterministic regression vs.\ flow-matching variants (5,000 steps, 24-frame
sequences). DF: diffusion forcing; HG: history guidance.}
\label{tab:regression_flow_matching}
\scriptsize
\setlength{\tabcolsep}{5pt}
\resizebox{\columnwidth}{!}{%
\begin{tabular}{lccc}
\toprule
Method & Occupancy IoU $\uparrow$ & 2D flow error $\downarrow$ & 3D flow error $\downarrow$ \\
\midrule
Binary occupancy target        & $0.330$ & $0.724$ & $0.733$ \\
Deterministic regression     & $0.208$ & $0.822$ & $0.762$ \\
Conditional (DF)             & $0.572$ & $0.700$ & $\best{0.680}$ \\
Vanilla guidance (DF + HG)   & $\best{0.709}$ & $\best{0.685}$ & $\best{0.655}$ \\
Fractional guidance (DF + HG)& $\secondbest{0.660}$ & $\secondbest{0.687}$ & $0.694$ \\
\bottomrule
\end{tabular}}
\end{table}

\begin{table}[h!]
    \centering
    \caption{
        Ablation of the occupancy and occlusion losses after 5,000
        training steps. We report occupancy IoU together with 2D and
        3D flow-estimation error.
    }
    \label{tab:occupancy_occlusion_loss_ablation}

    \small
    \setlength{\tabcolsep}{5pt}

    \begin{tabular}{lccc}
        \toprule
        Configuration
        & Occupancy IoU $\uparrow$
        & 2D flow error $\downarrow$
        & 3D flow error $\downarrow$ \\
        \midrule

        Full model
        & $\mathbf{0.709}$
        & $\mathbf{0.685}$
        & $\mathbf{0.655}$ \\

        Without occupancy loss
        & $0.572$
        & $0.854$
        & $\underline{0.713}$ \\

        Without occlusion loss
        & $\underline{0.580}$
        & $\underline{0.837}$
        & $0.804$ \\

        Without both losses
        & $0.572$
        & $0.946$
        & $0.829$ \\

        \bottomrule
    \end{tabular}
\end{table}

\section{Downstream tasks}
\subsection{Latent-to-video decoding}
We render predicted dynamics using a trainable convolutional mapper followed by a frozen pretrained image VAE decoder. The mapper translates spatial V-JEPA features into VAE latents; rendering is deterministic and does not involve a diffusion denoising model. V-JEPA and the VAE remain frozen during mapper training. 

\paragraph{Inputs and appearance conditioning.}
For camera view $i$, let $\hat z_{t+1}^{(i)}\in \mathbb{R}^{1408\times30\times30}$ denote the predicted feature map after projection into that view. Appearance conditioning uses only the first observed RGB frame of the corresponding view, resized to $240\times240$. Writing this image as $x_0^{(i)}\in[0,1]^{3\times240\times240}$, we compute a fixed anchor
\begin{equation}
 a^{(i)}=s\,\mu_E(2x_0^{(i)}-1),\qquad s=0.18215,
\end{equation}
where $\mu_E$ is the posterior mean of the frozen VAE encoder. We use the pretrained \texttt{stabilityai/sd-vae-ft-mse} AutoencoderKL. The anchor has shape $4\times30\times30$ and is reused at every time step. Camera parameters enter the upstream feature projection, rather than the mapper itself.

\paragraph{Mapper and RGB decoding.}
The mapper concatenates the features and appearance anchor along channels:
\begin{equation}
 \tilde y_{t+1}^{(i)}=
 g_\theta\!\left([\hat z_{t+1}^{(i)},a^{(i)}]\right).
\end{equation}
Its input therefore has 1412 channels. A $1\times1$ convolution maps these to 256 channels, followed by six residual blocks at unchanged spatial resolution. Each block applies GroupNorm with 32 groups, SiLU, a $3\times3$ convolution, GroupNorm, SiLU, and a second $3\times3$ convolution, with an identity skip connection. All convolutions in these blocks have 256 input and output channels, stride one, and padding one. The output head is GroupNorm, SiLU, and a $3\times3$ convolution producing four channels. The mapper has approximately 7.46 million trainable parameters and contains no temporal convolutions or attention layers. Its output is an absolute scaled VAE latent, not a residual added to the appearance anchor.

The frozen VAE decoder $D$ converts the mapped latent directly to RGB:
\begin{equation}
 \hat x_{t+1}^{(i)}=
 \operatorname{clip}_{[0,1]}\!\left(
 \frac{D(\tilde y_{t+1}^{(i)}/s)+1}{2}\right).
\end{equation}
Each latent state is decoded independently, yielding native $240\times240$ images. Larger exported videos are spatially resized from these images. Where the source frame rate exceeds the latent frame rate, our recent export pipeline linearly interpolates decoded RGB frames between latent timestamps and holds the last decoded frame through the remaining tail.

\paragraph{Mapper training.}
We train on paired cached V-JEPA maps and temporally aligned source RGB frames, resizing RGB targets to $240\times240$. The first-frame anchor is computed separately for each training view. Let $y_t=s\mu_E(2x_t-1)$, let $\Delta$ denote a consecutive-frame difference, and define $\rho(u,v)=\operatorname{mean}\sqrt{(u-v)^2+10^{-6}}$.
The selected base model uses
\begin{equation}
\begin{split}
 \mathcal L ={}& \rho(\tilde y,y)+\rho(\hat x,x)
 +0.5\,\operatorname{LPIPS}_{\mathrm{AlexNet}}(\hat x,x)\\
 &+0.5\,\rho(\Delta\hat x,\Delta x)
 +0.02\,\operatorname{TV}(\hat x),
\end{split}
\end{equation}
where TV is the sum of mean absolute horizontal and vertical image differences. LPIPS receives images normalized to $[-1,1]$. We use batches of six two-state clips, AdamW with learning rate $2\times10^{-4}$ and weight decay $10^{-4}$, gradient-norm clipping at 1, and bfloat16 automatic mixed precision. Image losses backpropagate through the frozen VAE decoder into the mapper. We select checkpoints by validation PSNR; the deployed base checkpoint is from zero-indexed epoch 8.

\paragraph{Dataset-specific adaptation.}
For the PhysGaia-adapted checkpoint, we fine-tune the mapper for five epochs using the first eight latent states of the stationary training views and their cached RGB previews resized to the decoder resolution. We retain the base objective, use batch size four and learning rate $5\times10^{-5}$, and select on the corresponding validation split. For the liquid-transfer example, a separate adaptation uses 15 other fluid scenes for training and three for validation, excluding the target scene from both. It runs 400 steps with batches of two two-state clips and learning rate $5\times10^{-5}$. Its objective is VAE-latent MSE plus RGB L1 plus $0.25$ times temporal-difference L1; selection uses validation RGB L1. Neither procedure trains a diffusion model, constructs a single-image augmentation bank, nor applies token dropout.

\subsection{Motion-guided video generation} \label{app:motion_guided_video}
We also use the predicted latent dynamics to guide video generation with Wan-Move~\cite{chu2025wan}. The pipeline has three steps: we estimate optical flow from consecutive latent maps, turn the flow into point trajectories,
and use these trajectories to guide Wan-Move. We give Wan-Move an initial RGB image, a text prompt, and the trajectory coordinates and visibility masks. 

\paragraph{Video features and latent-to-flow mapping.}\label{app:flow_decoder}
For real videos, we resize pairs of consecutive RGB frames to $480\times480$, apply ImageNet normalization, and encode each pair with frozen V-JEPA. This gives a latent map $z_k\in\mathbb{R}^{1408\times30\times30}$, which we assign to the second frame of the pair. 

The flow head first applies LayerNorm across the channels of each spatial token, without learned scale or bias. It then uses a shared $1\times1$ convolution to mix and reduce the 1408 channels to 48, followed by GELU and a $3\times3$ convolution with 48 channels. Let $a_k=P(z_k)$ denote these compressed features. The head concatenates $[a_k,a_{k+1},a_{k+1}-a_k]$, giving 144 channels, and applies a $3\times3$ convolution to 64 channels, eight-group GroupNorm, GELU, and two residual blocks. Each block applies a $3\times3$ convolution, eight-group GroupNorm, GELU, and another $3\times3$ convolution, then adds its input and applies GELU. A final $3\times3$ convolution produces 32 channels, which PixelShuffle with a factor of four rearranges into a two-channel $120\times120$ displacement field. All $3\times3$ convolutions use stride one and padding one. The selected checkpoints use only the two consecutive feature maps and their difference. The output values are in token units, so we multiply them by four to express displacement in pixels on the flow grid. Each displacement covers the interval between two latent timestamps.

\paragraph{Optical reference and trajectory integration.}
We use OpenCV DIS to estimate optical flow from adjacent RGB frames as an apparent-motion reference. We select moving points across a $30\times30$ grid using forward--backward consistency and image evidence, and keep at most 150 tracks for the sparse videos. All compared methods start from the same points. We build trajectories by sampling each flow field at the current point location:
\begin{equation}
 p_{k+1}=p_k+\operatorname{bilinear}(u_k,p_k).
\end{equation}
We scale coordinates consistently between the flow grid and image, interpolate positions between latent timestamps, and stop tracks that leave the image or lose valid support.

\paragraph{Wan-Move conditioning and generation.}
We give Wan-Move the initial RGB image, a short essential text prompt for it to understand the scene, and the selected trajectories with visibility masks. We keep only tracks that are valid at the start of the clip and resample them to 81 timestamps. For the real-video examples, this stretches the selected motion over the output duration. The multiview pipeline also supports preserving the original timing and holding the final positions for the remaining frames.

We use the pretrained Wan-Move 14B 480P checkpoint without further fine-tuning, generating 81 frames at 16 fps with 40 UniPC steps, schedule shift 3, guidance scale 5, and seed 42. Output dimensions follow the input aspect ratio within a $480\times832$ pixel area budget. These videos show how the estimated motion can guide generation; they do not directly measure the accuracy of the predictor or flow head.

\paragraph{Results}\label{app:real_vids}
The video results are shown in \Cref{fig:real_vid}. See supplemental webpage for accompanying videos. 
\begingroup

\setlength{\framesetwidth}{\linewidth}
\renewcommand{\LTcaptype}{figure}

\setlength{\LTcapwidth}{\framesetwidth}
\captionsetup{position=top}

\begin{longtable}{@{}c@{}}
    \caption{Input frames and the corresponding generated sequences.}\label{fig:real_vid}
    \label{fig:input_generated} \\[8pt]

    \framesetheader \\[8pt]
    \endfirsthead

    \framesetheader \\[8pt]
    \endhead
    \endfoot
    \endlastfoot

    \frameset{Example 1: Fruits drop in water}
    {

        \framecell{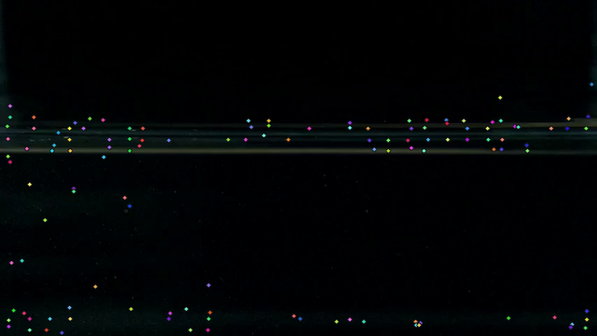}{1} &
        \framecell{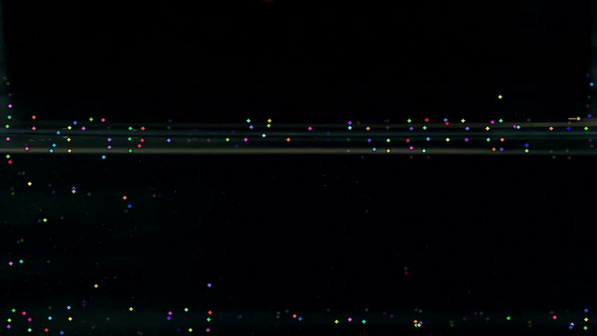}{2} \\[6pt]

        \framecell{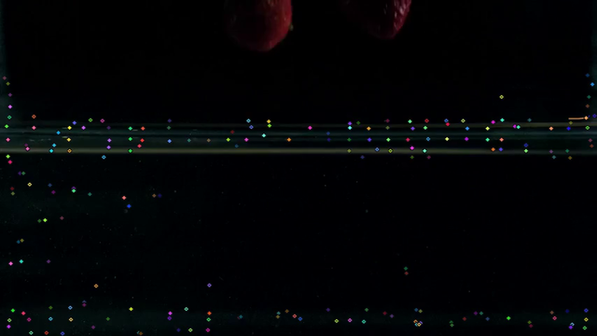}{3} &
        \framecell{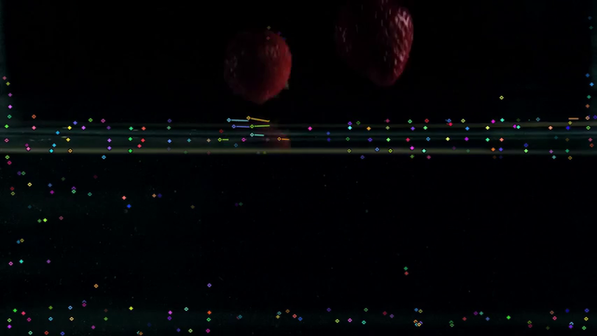}{4}
    }
    {

        \framecell{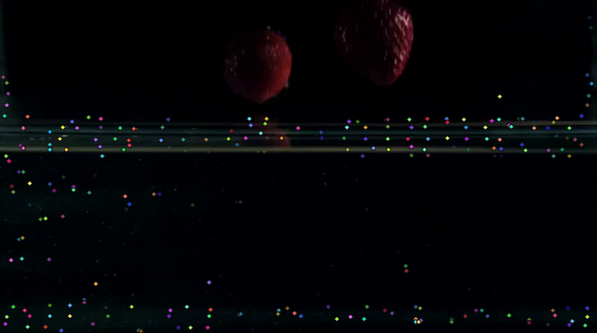}{5} &
        \framecell{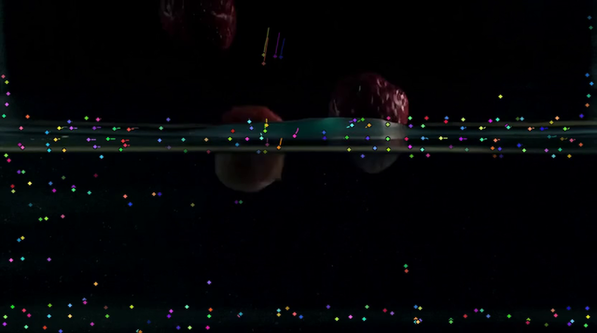}{6} &
        \framecell{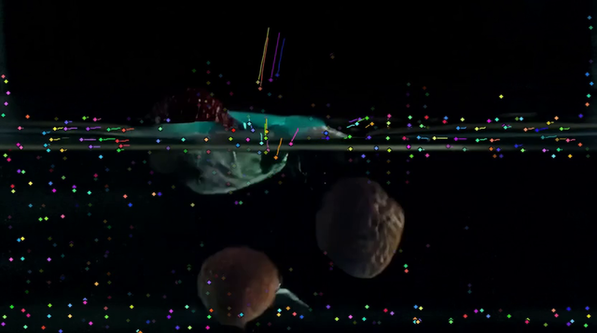}{7} &
        \framecell{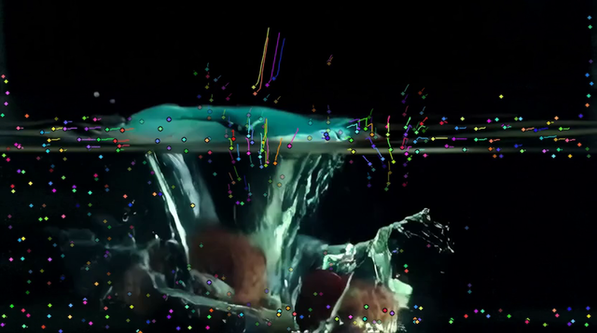}{8} \\[6pt]

        \framecell{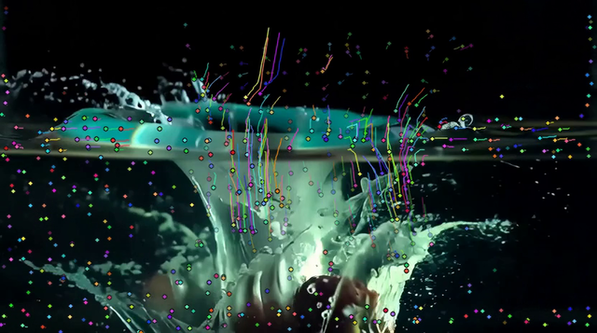}{9} &
        \framecell{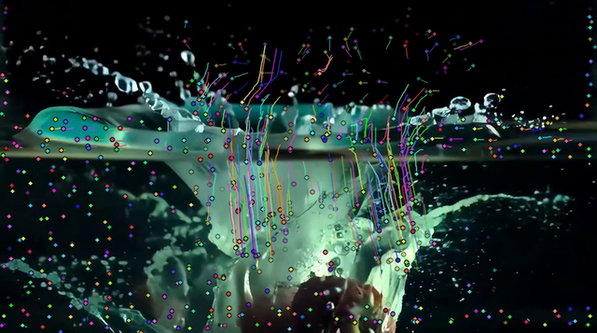}{10} &
        \framecell{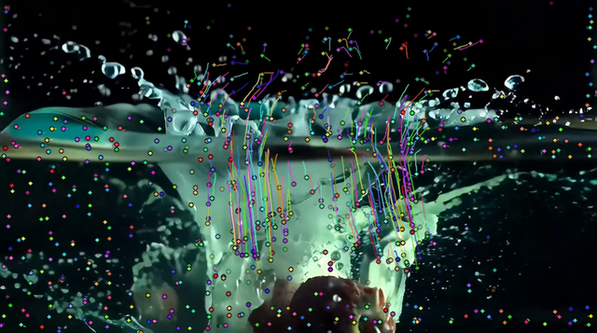}{11} &
        \framecell{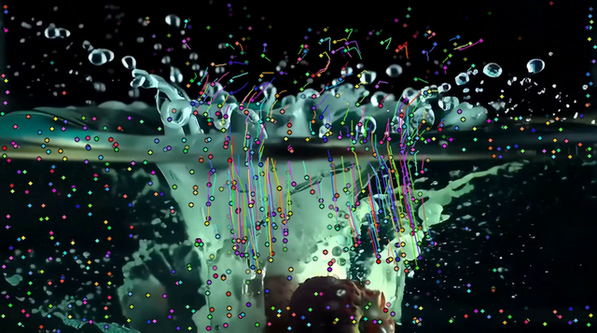}{12} 
    }
    \\
    
    \frameset{Example 2: Coffee pouring}
    {

        \framecell{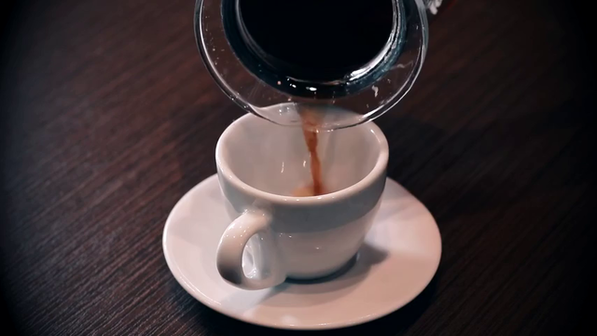}{1} &
        \framecell{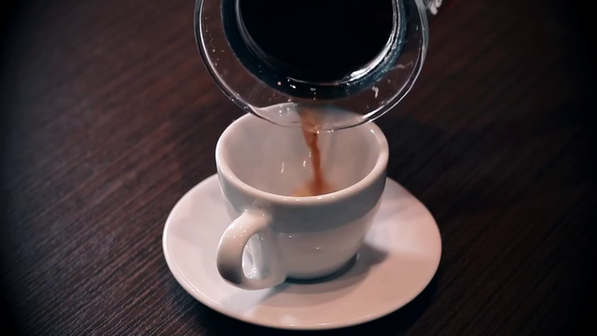}{2} \\[6pt]

        \framecell{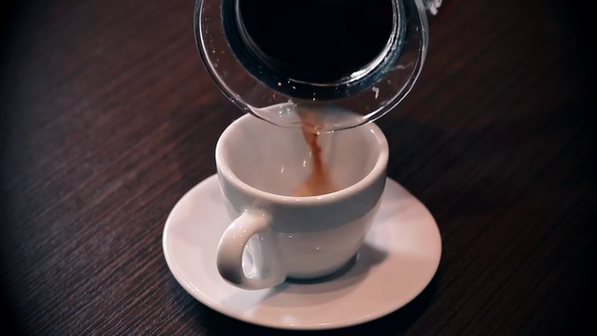}{3} &
        \framecell{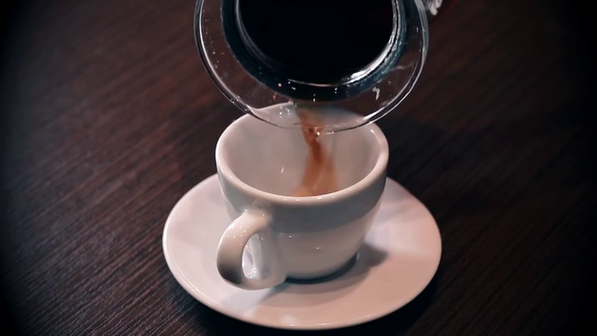}{4}
    }
    {

        \framecell{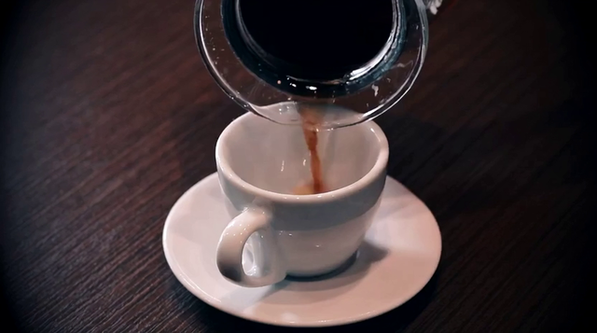}{5} &
        \framecell{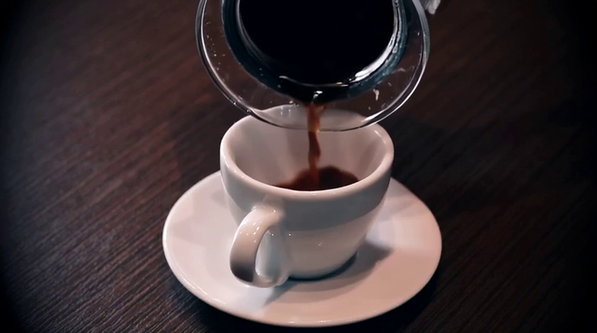}{6} &
        \framecell{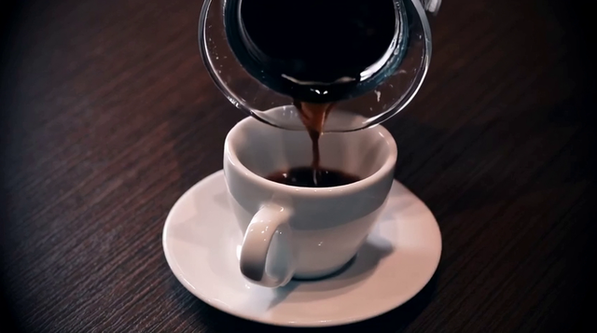}{7} &
        \framecell{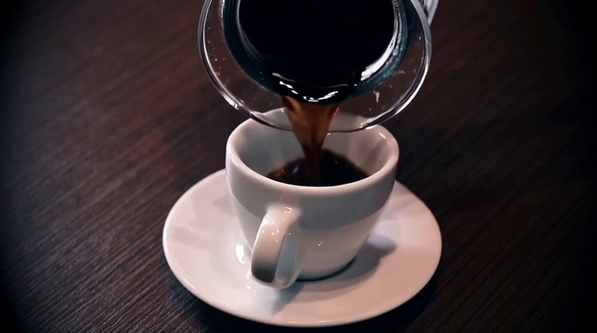}{8} \\[6pt]

        \framecell{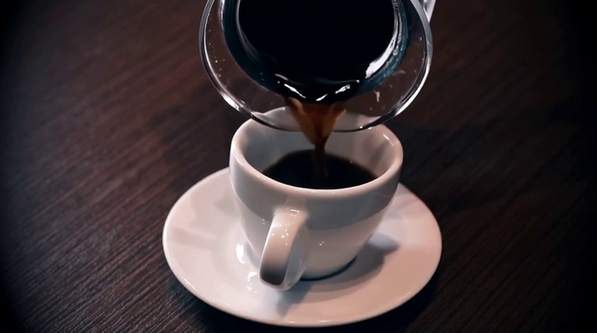}{9} &
        \framecell{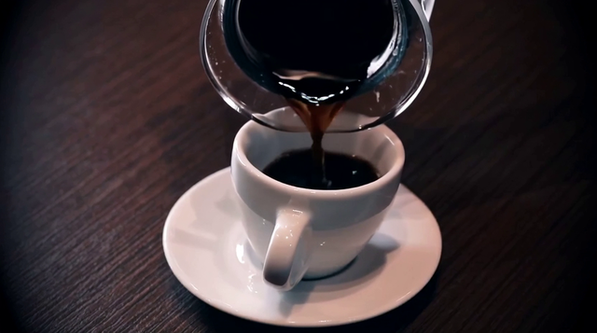}{10} &
        \framecell{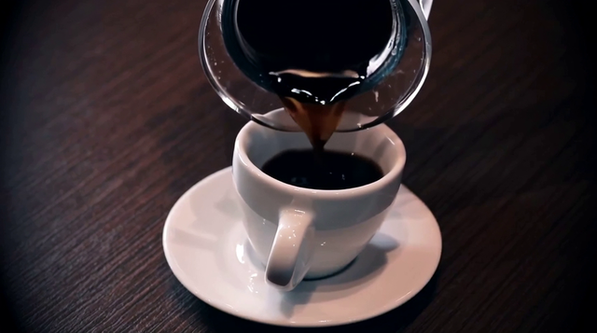}{11} &
        \framecell{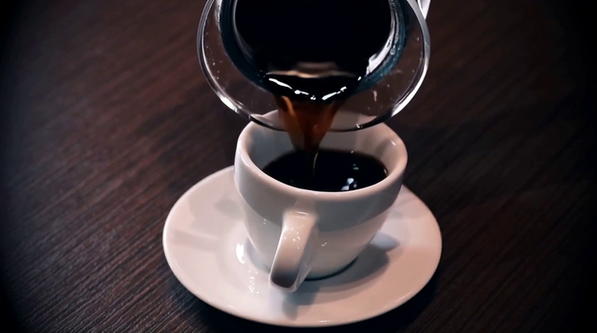}{12} 
    }
    \\

    \frameset{Example 3: Paper on fire}
    {

        \framecell{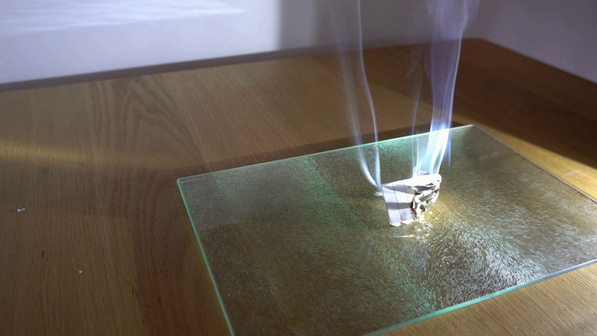}{1} &
        \framecell{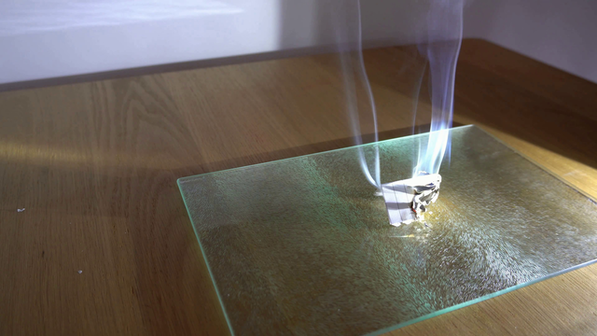}{2} \\[6pt]

        \framecell{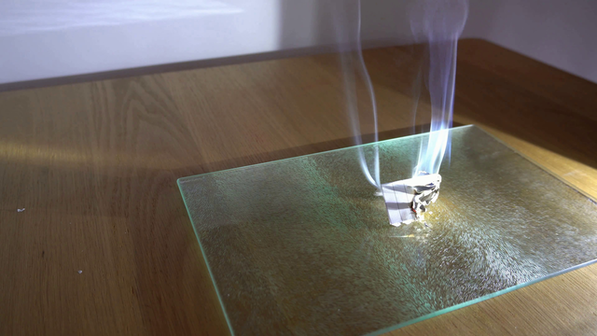}{3} &
        \framecell{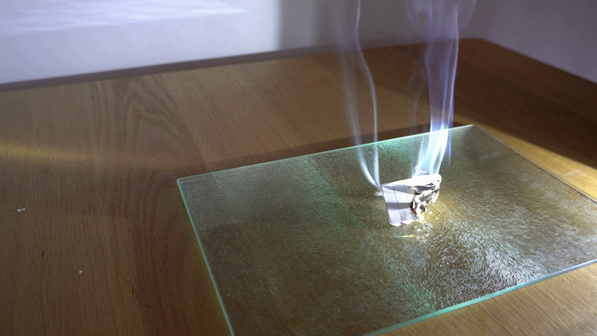}{4}
    }
    {

        \framecell{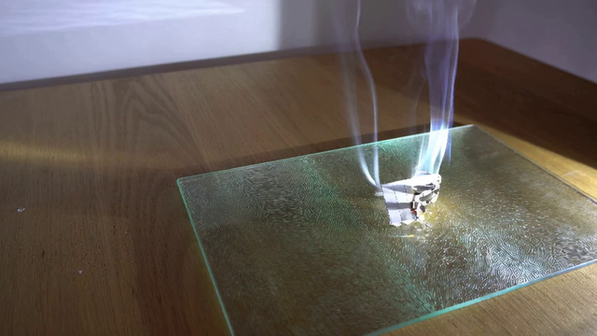}{5} &
        \framecell{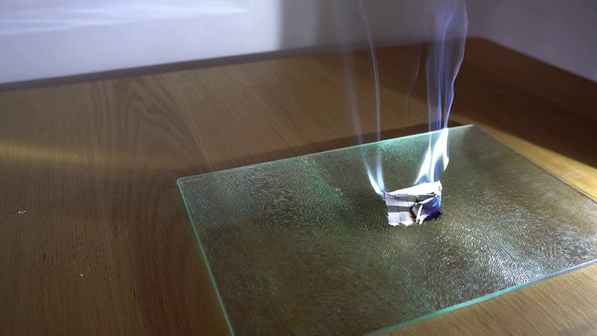}{6} &
        \framecell{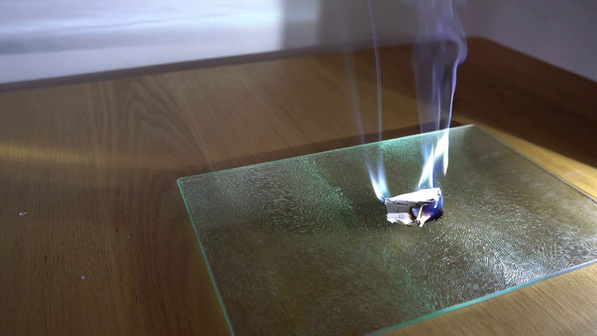}{7} &
        \framecell{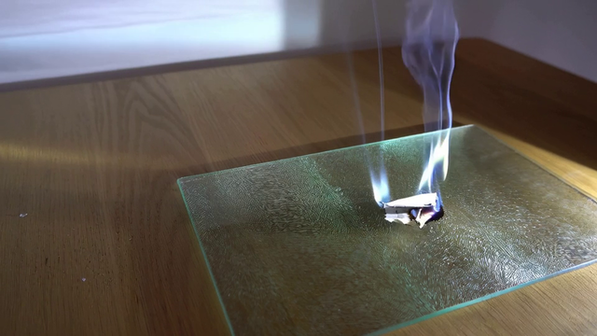}{8} \\[6pt]

        \framecell{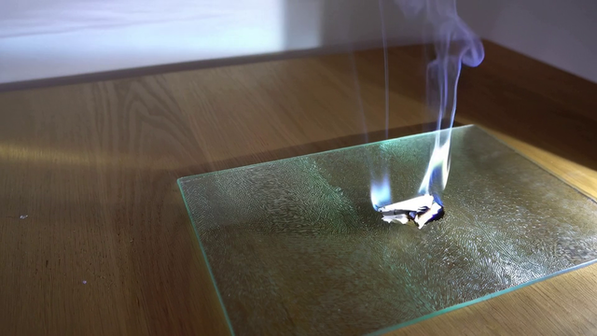}{9} &
        \framecell{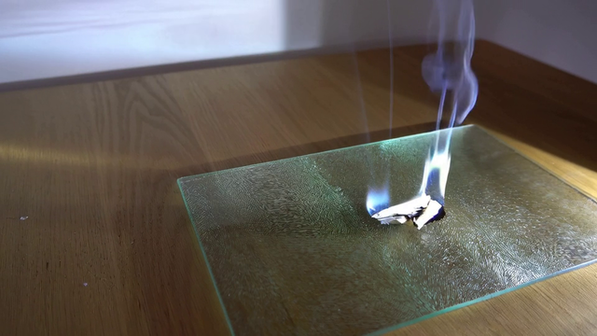}{10} &
        \framecell{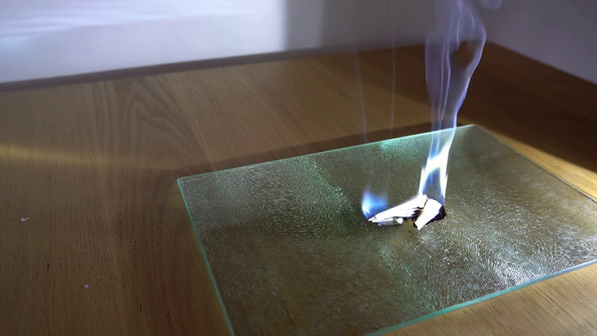}{11} &
        \framecell{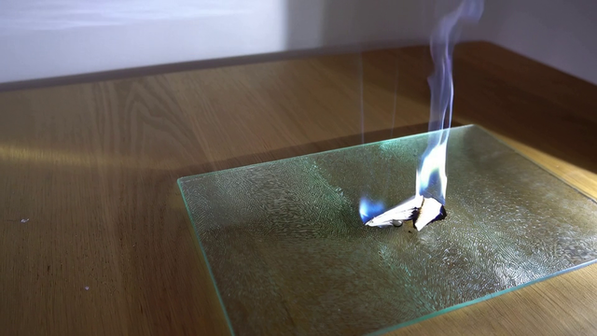}{12} 
    }
    \\

    \frameset{Example 4: Liquid with duck}
    {

        \framecell{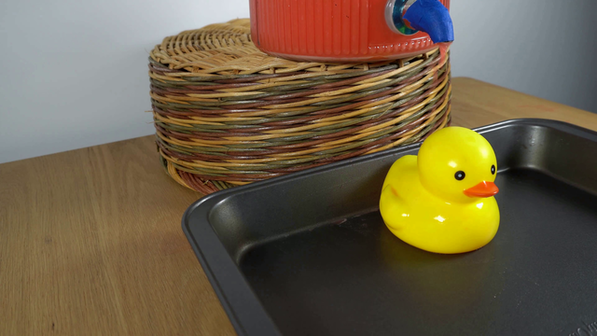}{1} &
        \framecell{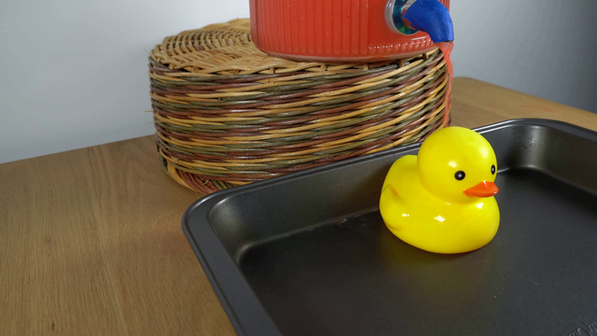}{2} \\[6pt]

        \framecell{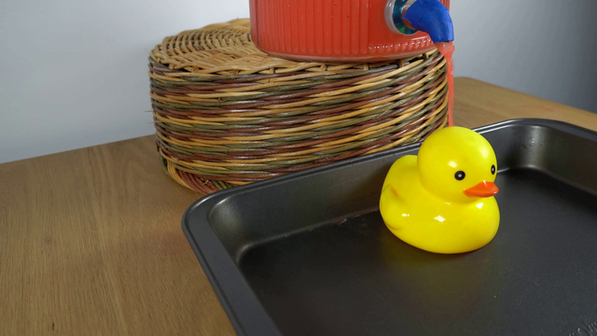}{3} &
        \framecell{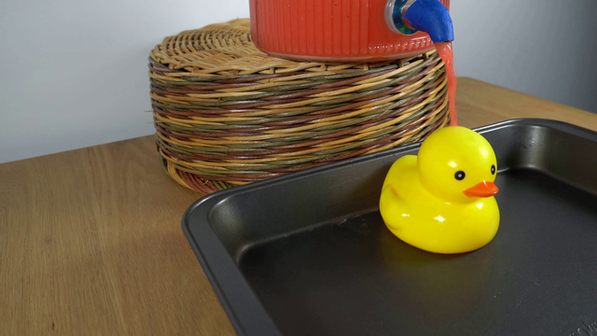}{4}
    }
    {

        \framecell{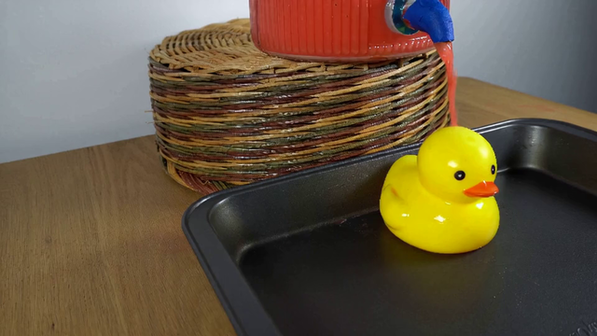}{5} &
        \framecell{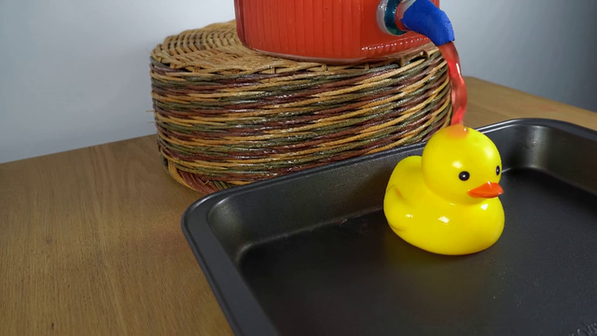}{6} &
        \framecell{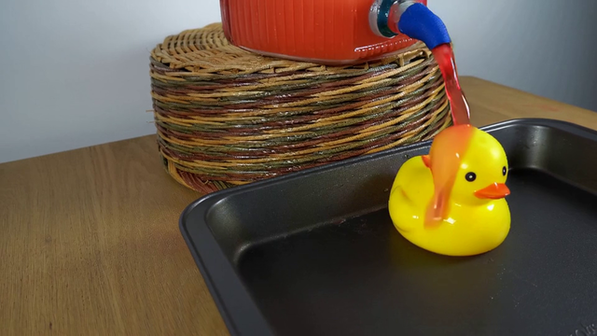}{7} &
        \framecell{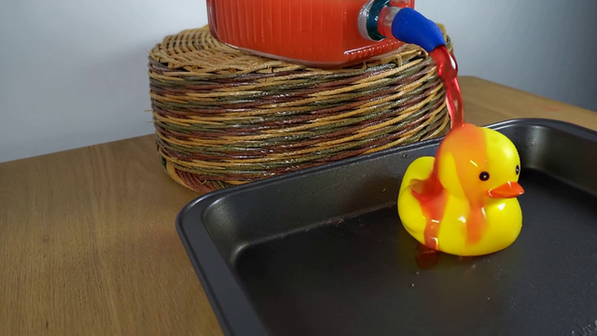}{8} \\[6pt]

        \framecell{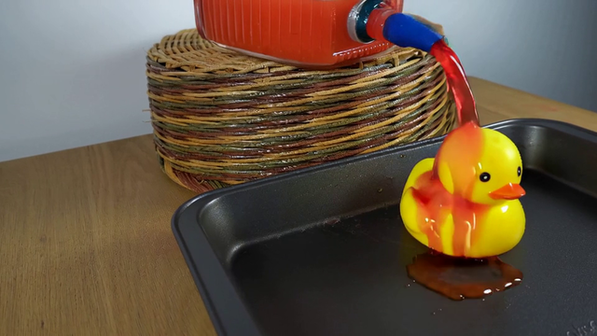}{9} &
        \framecell{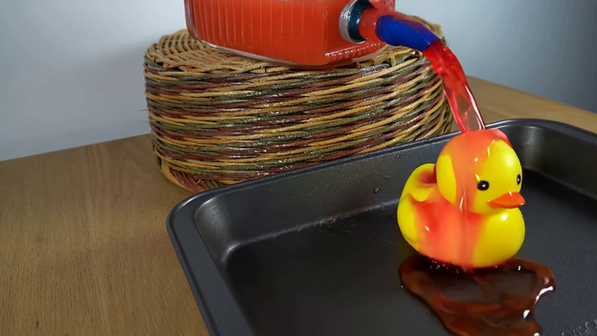}{10} &
        \framecell{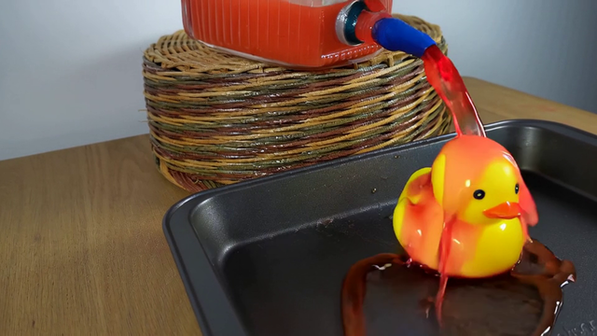}{11} &
        \framecell{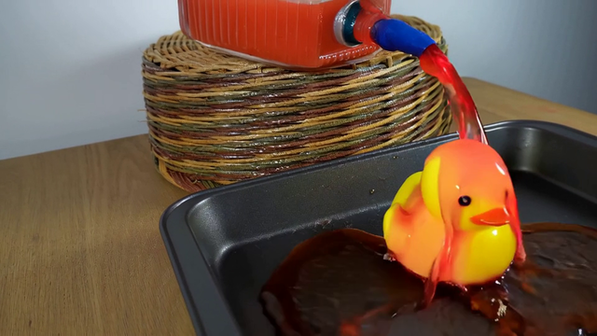}{12} 
    }
    \\

    \frameset{Example 5: Potato in water}
    {

        \framecell{figures/wan-move_video/0144_perspective-right_trimmed-potato-in-water/0144_perspective-right_trimmed-potato-in-water_in_0.png}{1} &
        \framecell{figures/wan-move_video/0144_perspective-right_trimmed-potato-in-water/0144_perspective-right_trimmed-potato-in-water_in_1.png}{2} \\[6pt]

        \framecell{figures/wan-move_video/0144_perspective-right_trimmed-potato-in-water/0144_perspective-right_trimmed-potato-in-water_in_2.png}{3} &
        \framecell{figures/wan-move_video/0144_perspective-right_trimmed-potato-in-water/0144_perspective-right_trimmed-potato-in-water_in_3.png}{4}
    }
    {

        \framecell{figures/wan-move_video/0144_perspective-right_trimmed-potato-in-water/0144_perspective-right_trimmed-potato-in-water_0.png}{5} &
        \framecell{figures/wan-move_video/0144_perspective-right_trimmed-potato-in-water/0144_perspective-right_trimmed-potato-in-water_4.png}{6} &
        \framecell{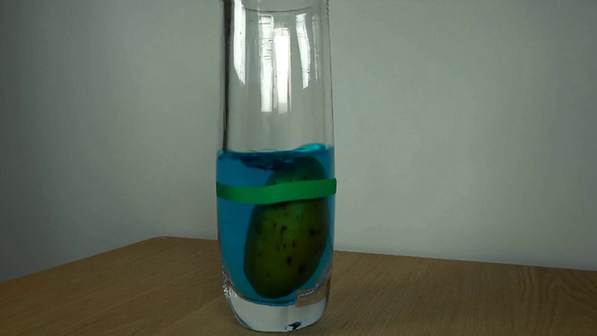}{7} &
        \framecell{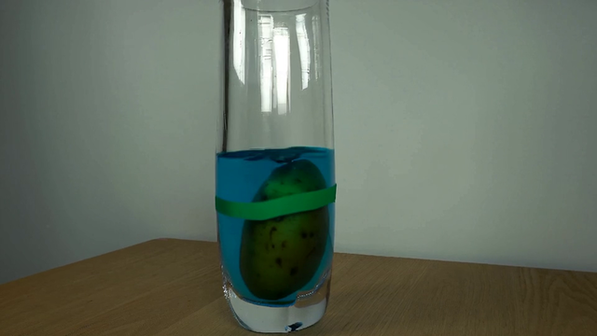}{8} \\[6pt]

        \framecell{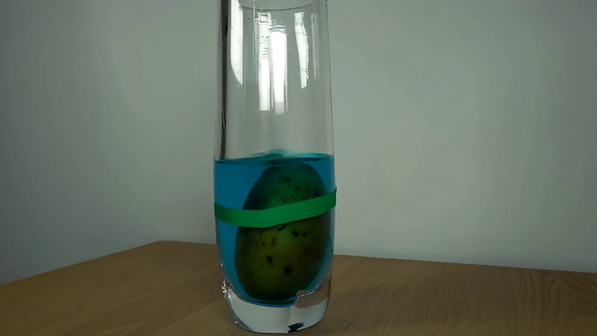}{9} &
        \framecell{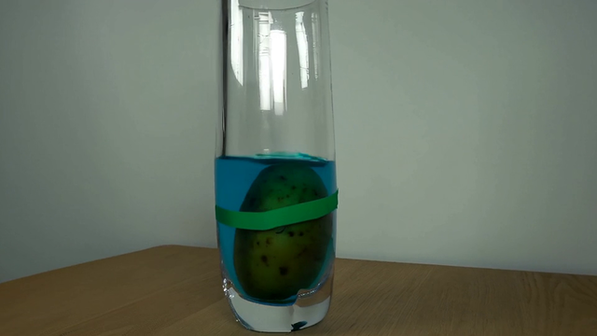}{10} &
        \framecell{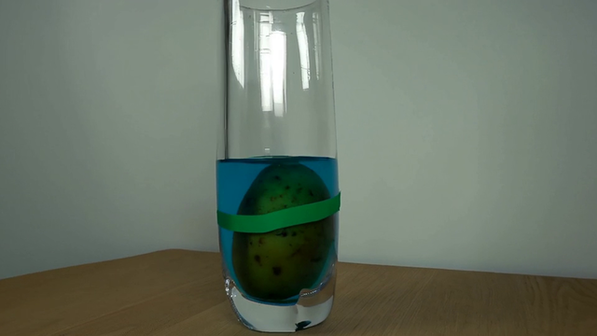}{11} &
        \framecell{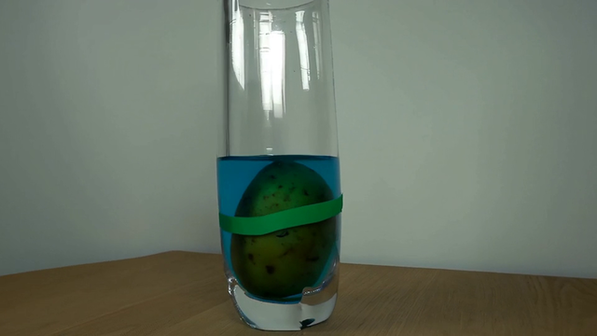}{12} 
    }
    \\

    \frameset{Example 6: Weight on pillow}
    {

        \framecell{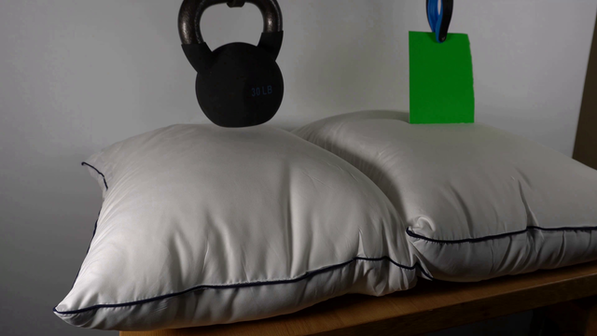}{1} &
        \framecell{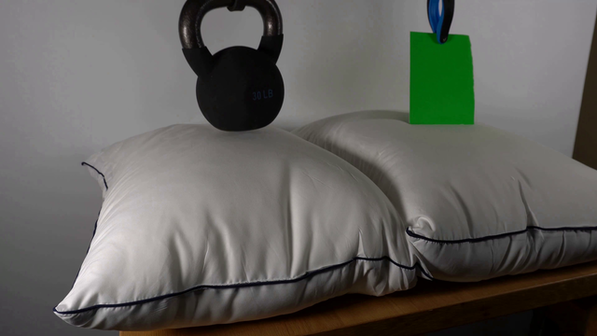}{2} \\[6pt]

        \framecell{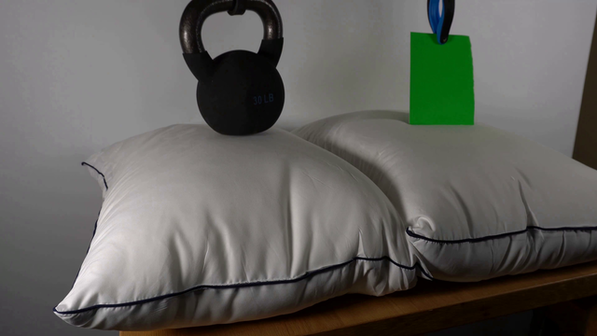}{3} &
        \framecell{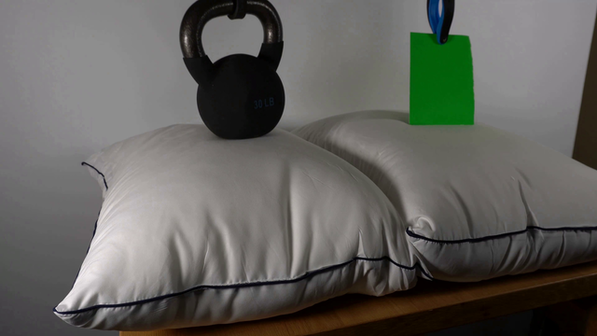}{4}
    }
    {

        \framecell{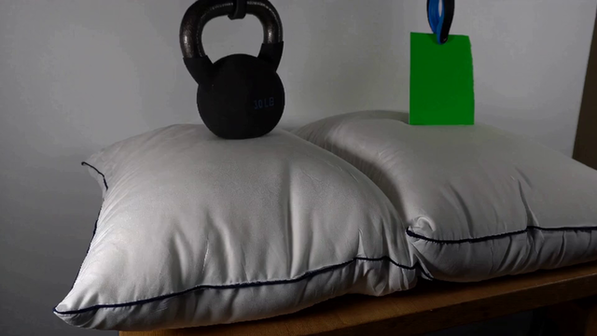}{5} &
        \framecell{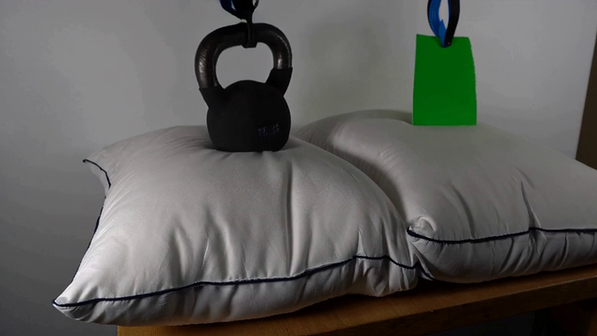}{6} &
        \framecell{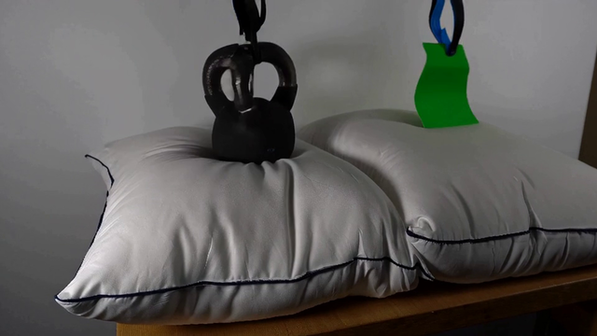}{7} &
        \framecell{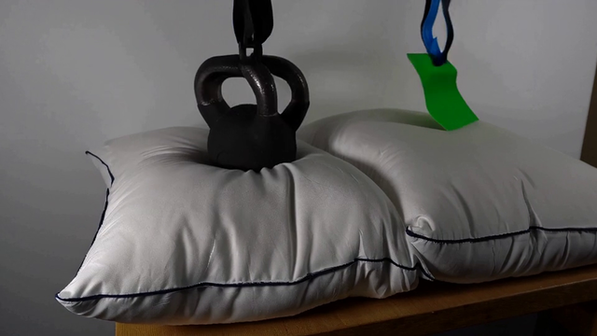}{8} \\[6pt]

        \framecell{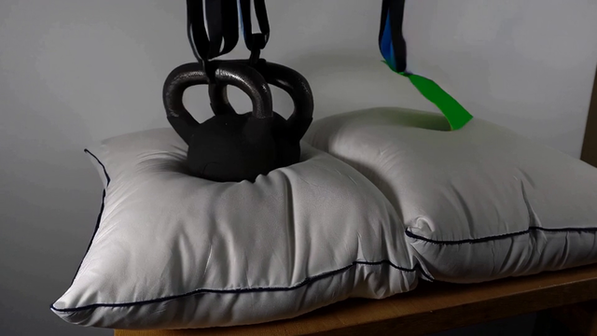}{9} &
        \framecell{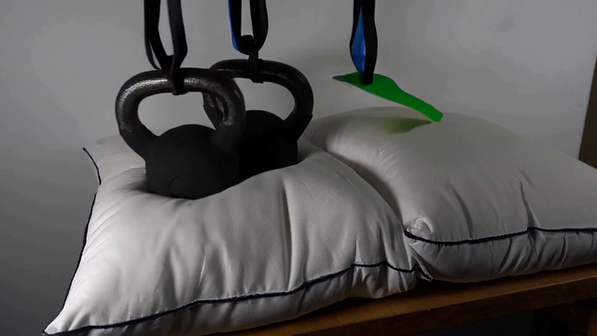}{10} &
        \framecell{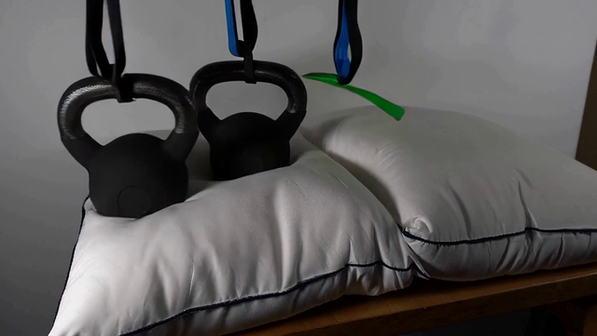}{11} &
        \framecell{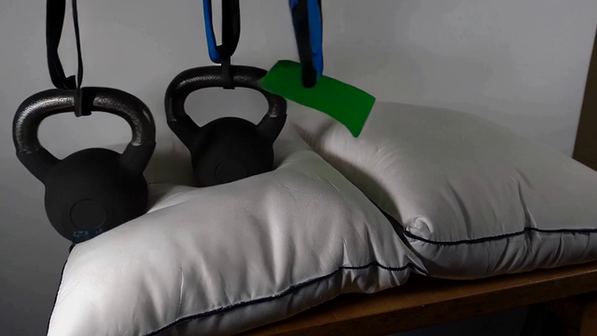}{12} 
    }
    \\

    \frameset{Example 7: Stone drop in water}
    {

        \framecell{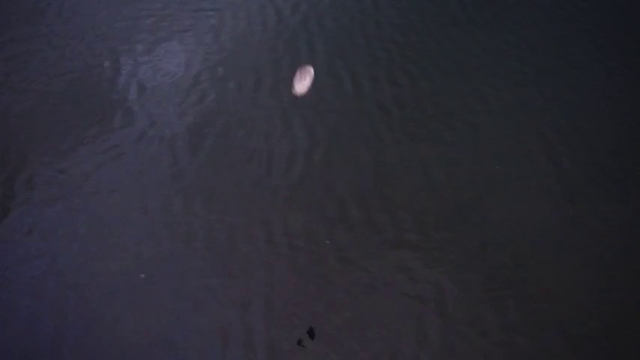}{1} &
        \framecell{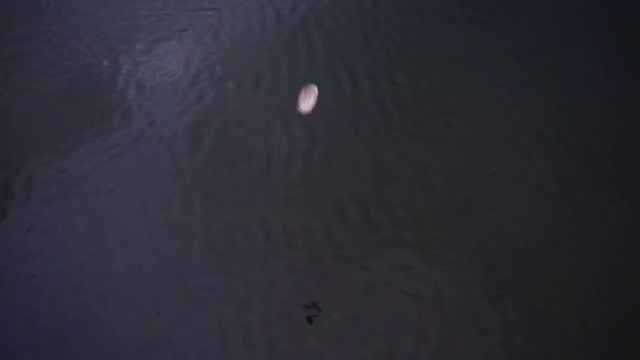}{2} \\[6pt]

        \framecell{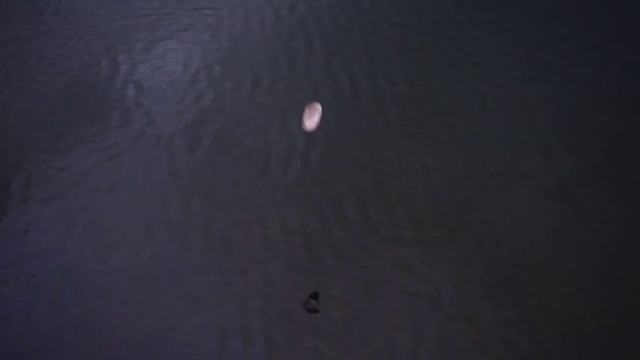}{3} &
        \framecell{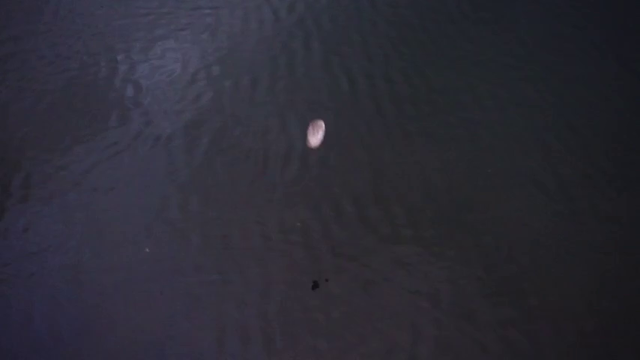}{4}
    }
    {

        \framecell{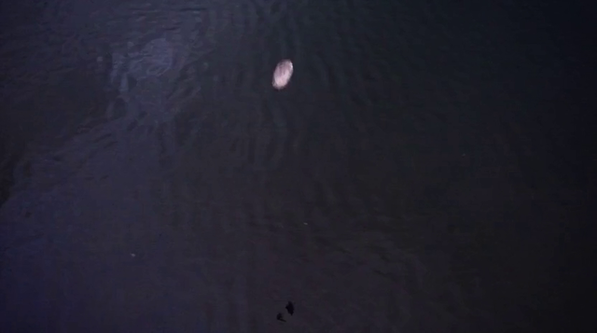}{5} &
        \framecell{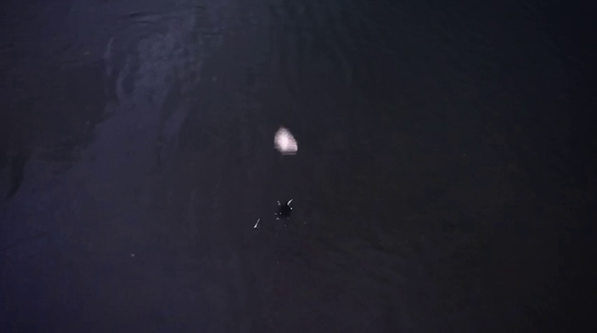}{6} &
        \framecell{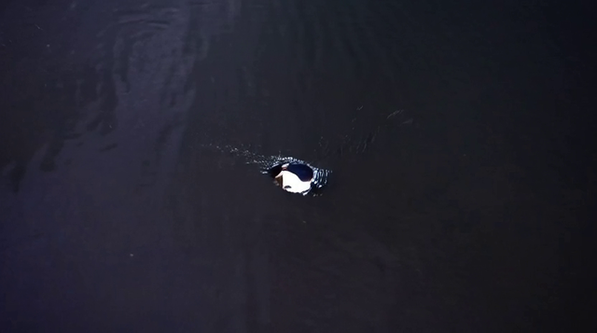}{7} &
        \framecell{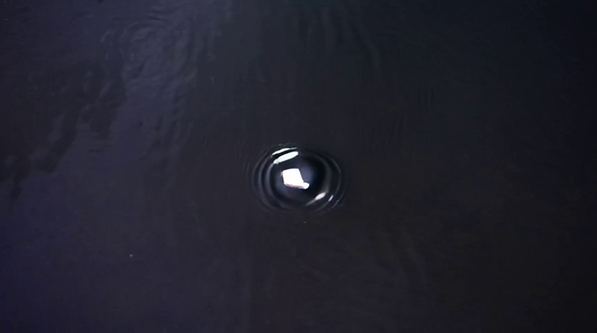}{8} \\[6pt]

        \framecell{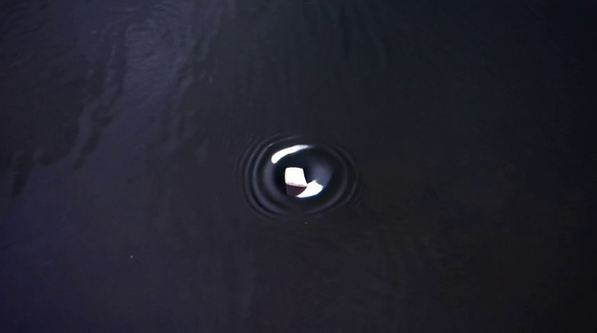}{9} &
        \framecell{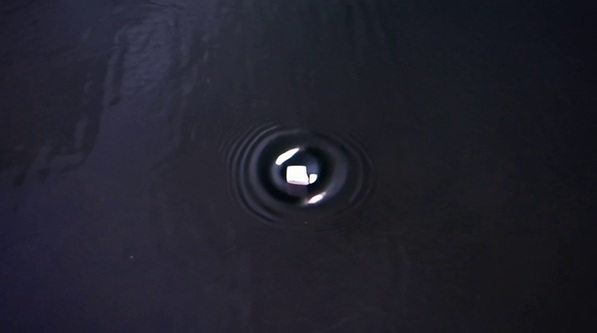}{10} &
        \framecell{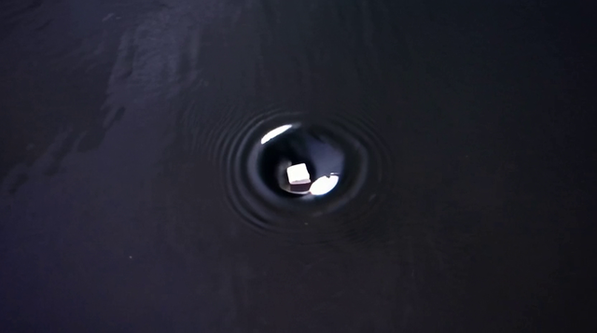}{11} &
        \framecell{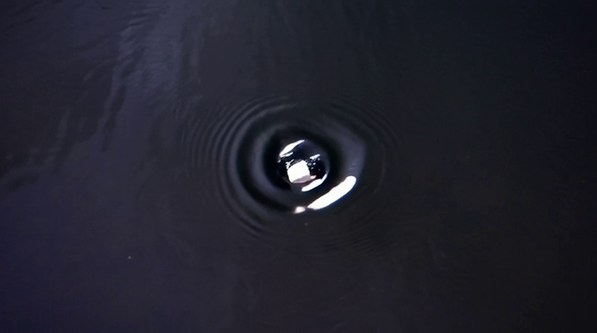}{12} 
    }
    \\

    \frameset{Example 8: Volcano}
    {

        \framecell{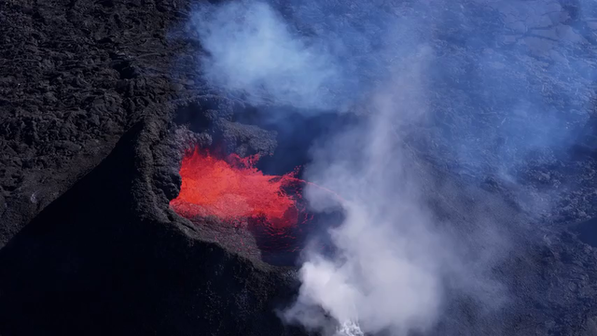}{1} &
        \framecell{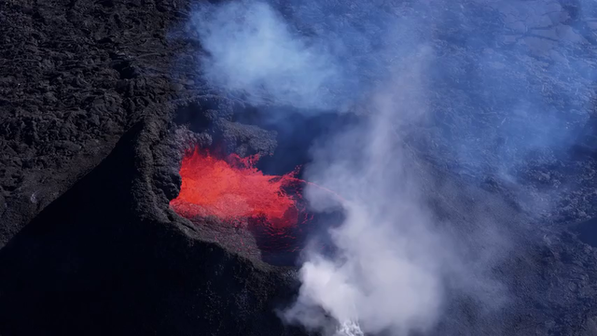}{2} \\[6pt]

        \framecell{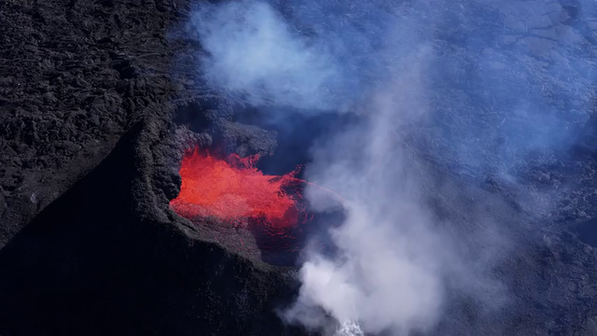}{3} &
        \framecell{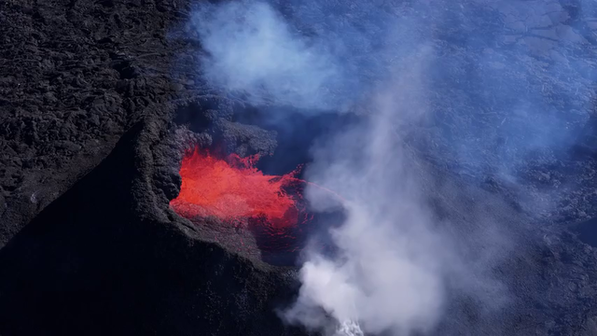}{4}
    }
    {

        \framecell{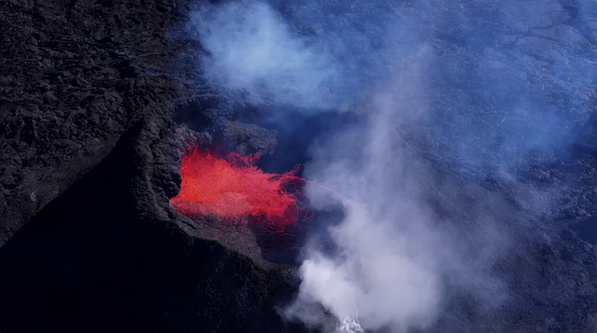}{5} &
        \framecell{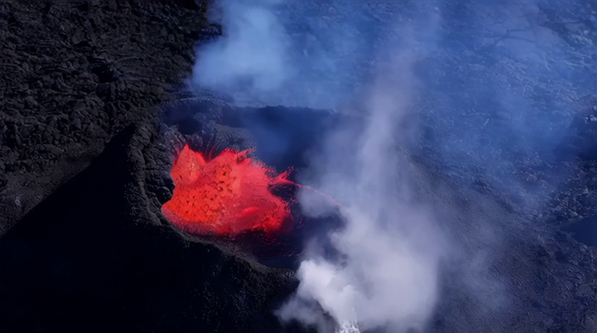}{6} &
        \framecell{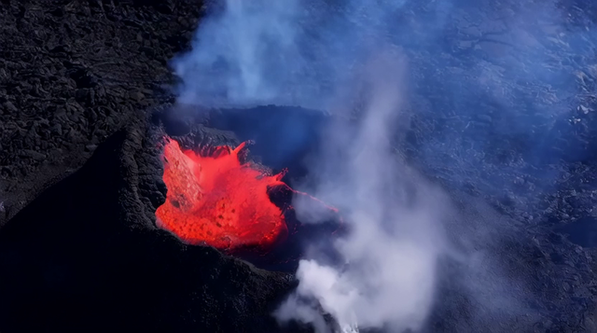}{7} &
        \framecell{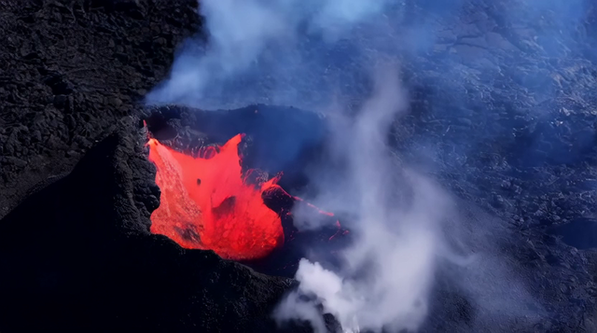}{8} \\[6pt]

        \framecell{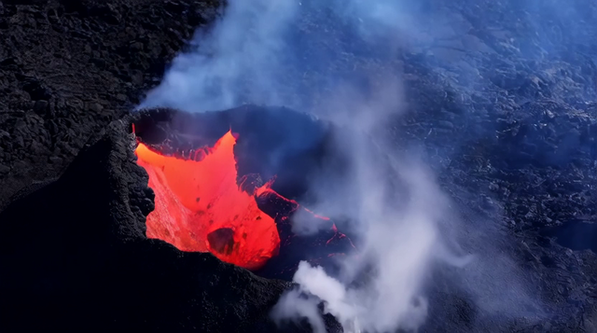}{9} &
        \framecell{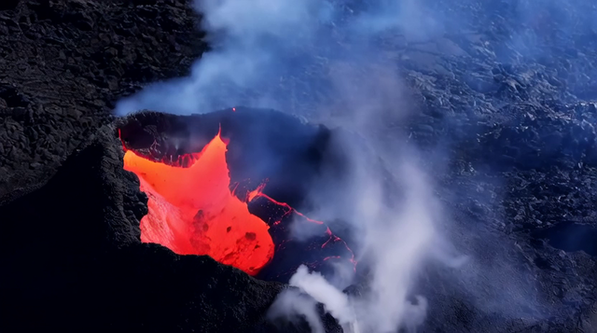}{10} &
        \framecell{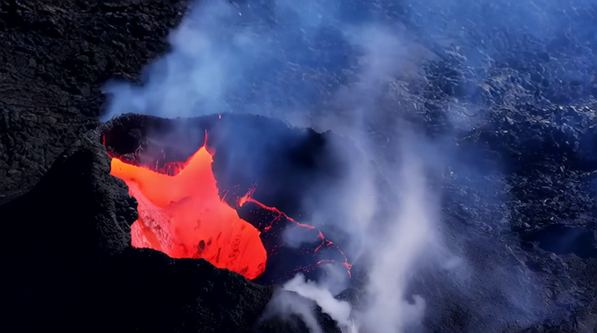}{11} &
        \framecell{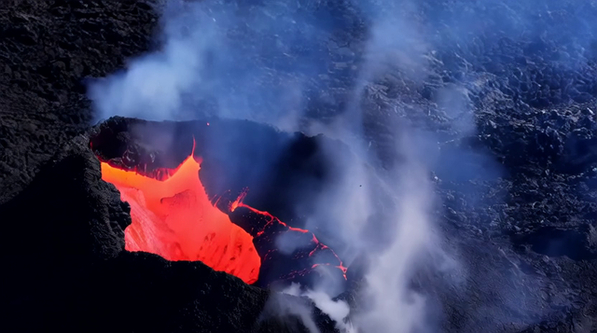}{12} 
    }
    \\

    \frameset{Example 9: Water drops}
    {

        \framecell{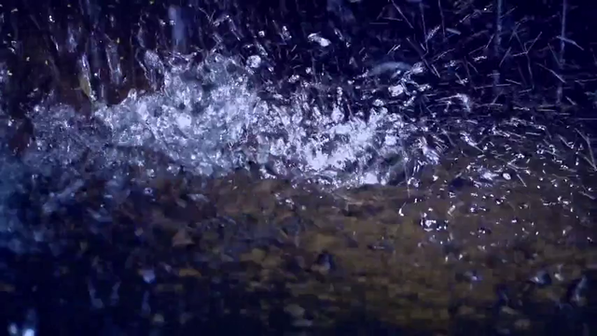}{1} &
        \framecell{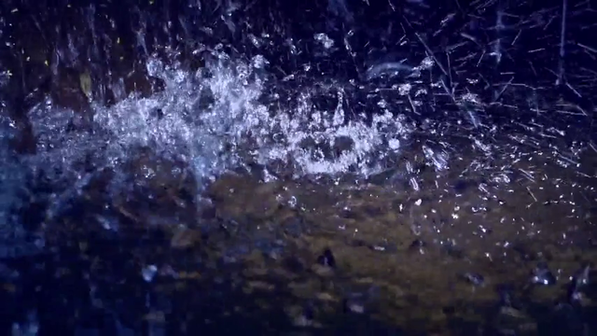}{2} \\[6pt]

        \framecell{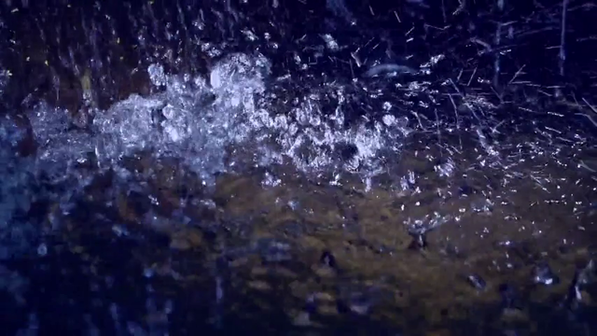}{3} &
        \framecell{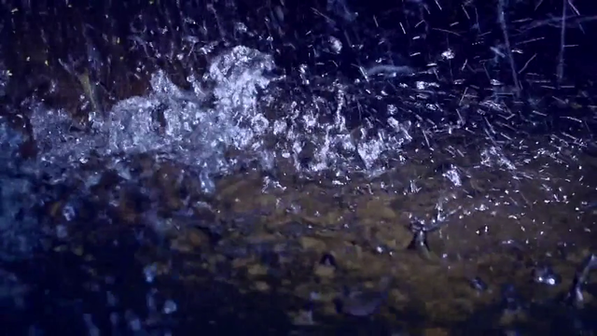}{4}
    }
    {

        \framecell{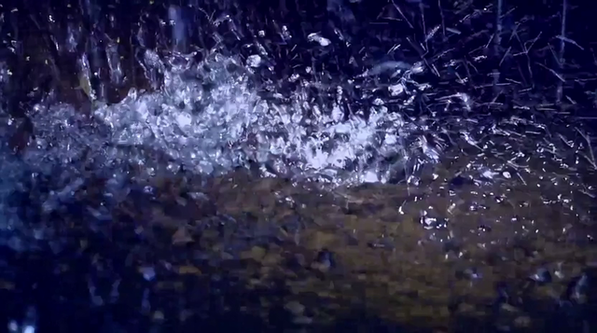}{5} &
        \framecell{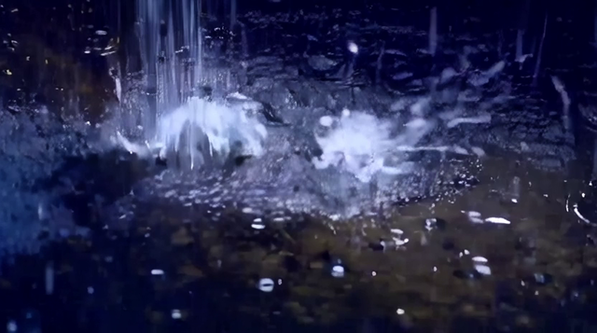}{6} &
        \framecell{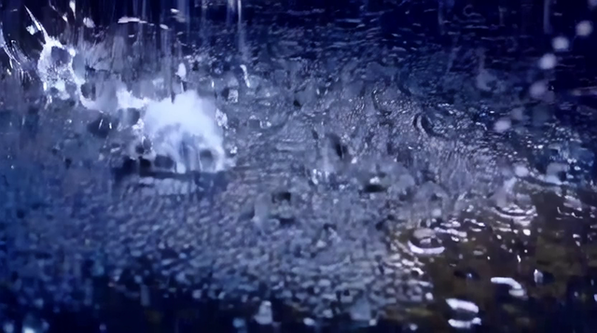}{7} &
        \framecell{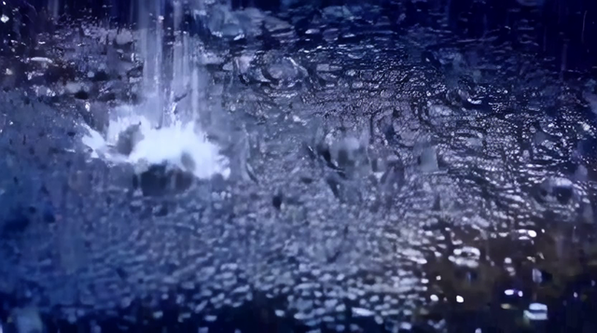}{8} \\[6pt]

        \framecell{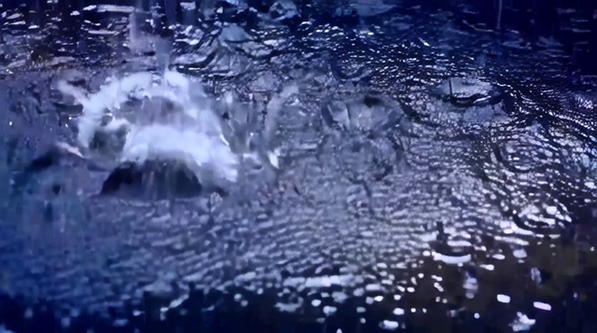}{9} &
        \framecell{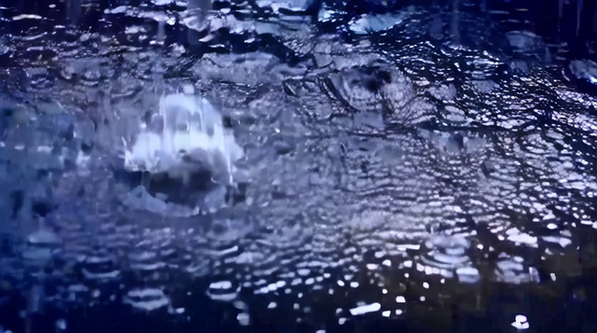}{10} &
        \framecell{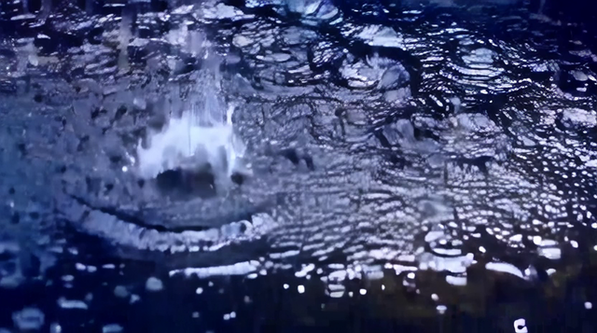}{11} &
        \framecell{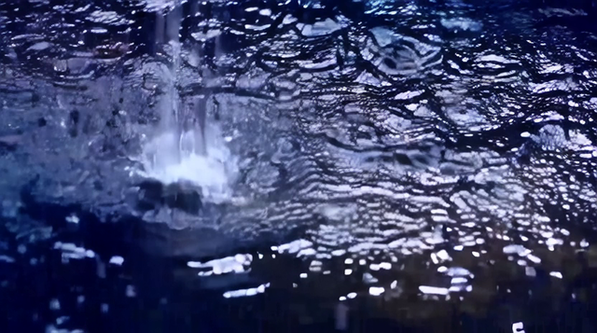}{12} 
    }
    \\

    \frameset{Example 10: Water waves}
    {

        \framecell{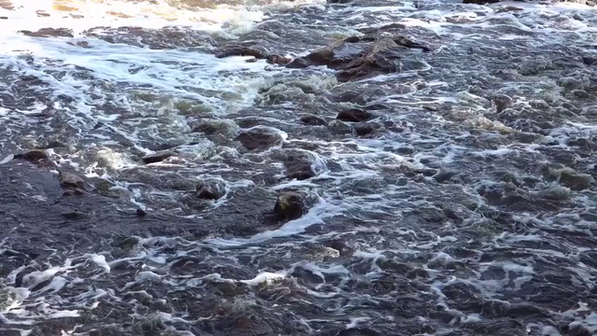}{1} &
        \framecell{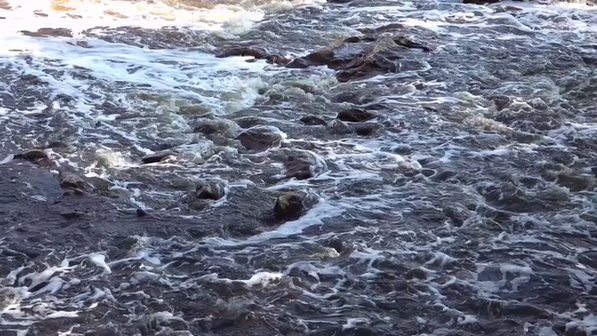}{2} \\[6pt]

        \framecell{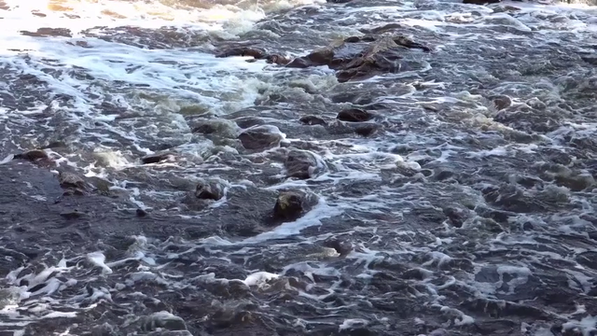}{3} &
        \framecell{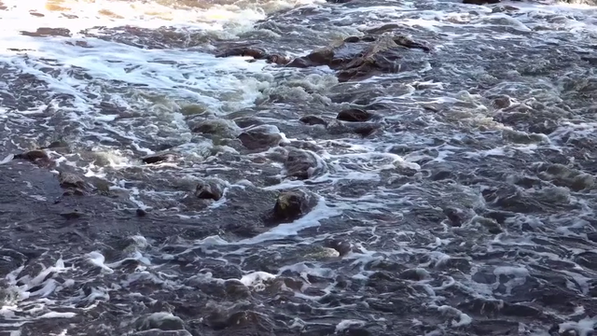}{4}
    }
    {

        \framecell{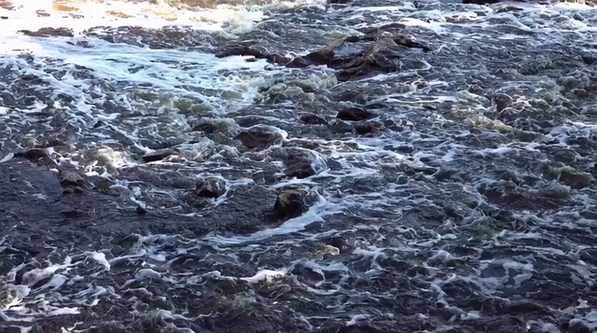}{5} &
        \framecell{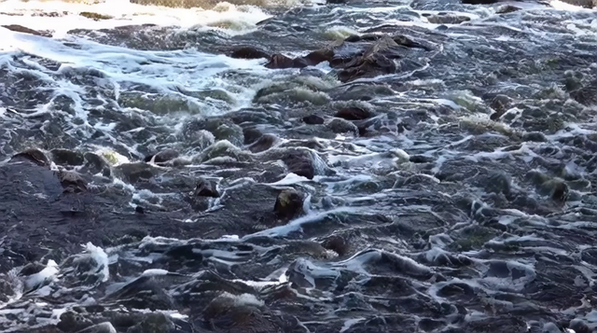}{6} &
        \framecell{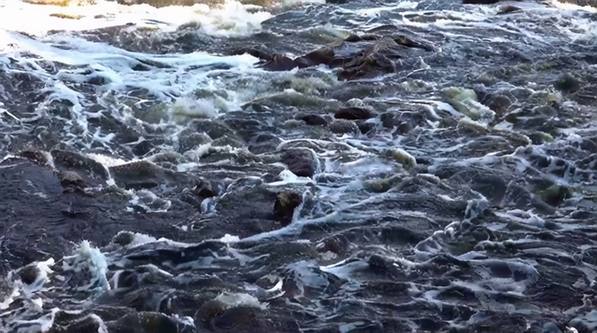}{7} &
        \framecell{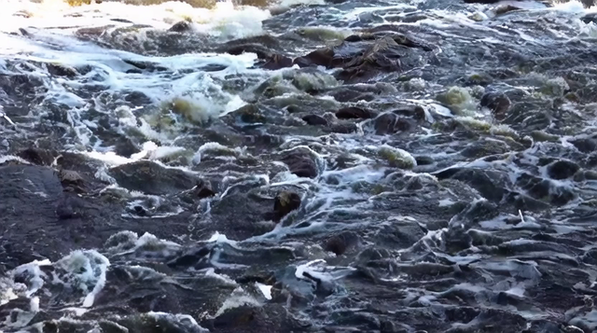}{8} \\[6pt]

        \framecell{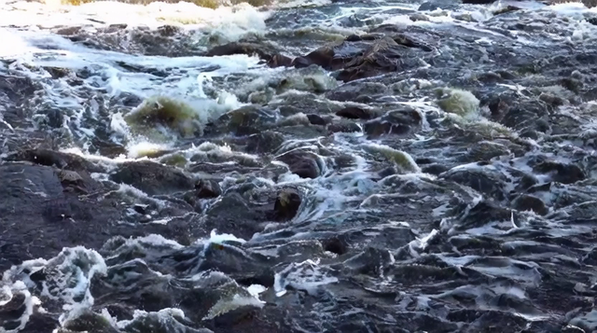}{9} &
        \framecell{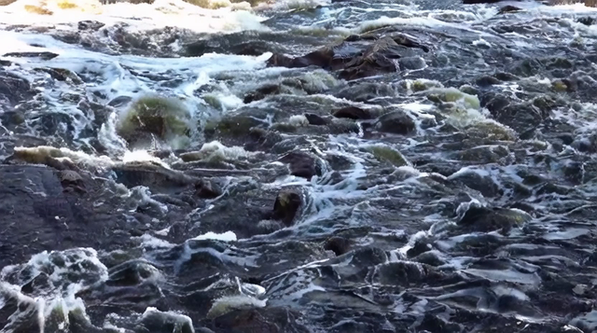}{10} &
        \framecell{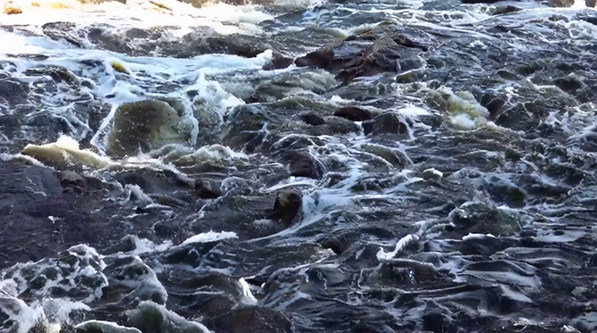}{11} &
        \framecell{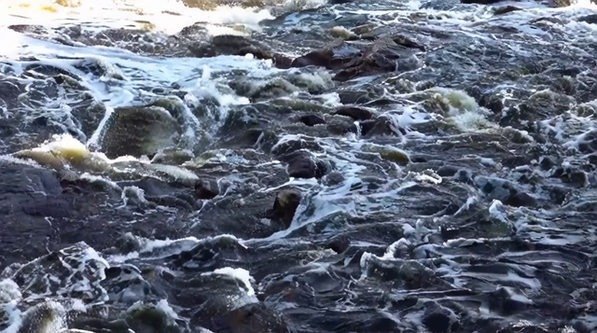}{12} 
    }
    \\

\end{longtable}

\endgroup

\subsection{Joint training ability}
Although our model is trained modularly, we can also adopt joint embedding learning, where the encoder latent is jointly trained with the predictor. However, this would require more data to generalize and reduce collapse, and would require much more compute. We hope that this is also one possible direction for future work.

\subsection{Broader societal impact}  
Grounding generative video models in Euclidean space holds significant positive potential for physical planning, autonomous driving, and robotics, where structural and physical accuracy are paramount for safety. By simulating realistic physical outcomes from novel views, our method could accelerate safe, offline reinforcement learning. However, as with any high-fidelity video generation, there is an inherent risk of misuse. Enhancing the physical realism of generated videos ironically increases their believability, potentially facilitating the creation of convincing misinformation or sophisticated deepfakes. Mitigating these risks will require the parallel development of robust video-forensic tools and watermarking techniques.

\end{document}